\def\arxiv % please commit this line before submit
\def\year{2020}\relax
\documentclass[letterpaper]{article} % DO NOT CHANGE THIS

\usepackage{aaai20}  % DO NOT CHANGE THIS
\usepackage{times}  % DO NOT CHANGE THIS
\usepackage{helvet} % DO NOT CHANGE THIS
\usepackage{courier}  % DO NOT CHANGE THIS
\usepackage[hyphens]{url}  % DO NOT CHANGE THIS
\usepackage{graphicx} % DO NOT CHANGE THIS
\usepackage{subfigure}
\usepackage{caption} % DO NOT CHANGE THIS AND DO NOT ADD ANY OPTIONS TO IT
\usepackage{amsmath}
\usepackage{amssymb}
\usepackage{xcolor}
\usepackage{multirow}

\newtheorem{lemma}{Lemma}

\ifdefined\arxiv
\nocopyright
\addtocounter{page}{2}
\usepackage[sort&compress,numbers]{natbib}
\else
\usepackage[authoryear]{natbib}
\fi
\renewcommand{\cite}{\citep}
\newcommand{\papertitle}{DARTS+: Improved Differentiable Architecture Search with Early Stopping}

\title{\papertitle}
\author{Paper ID: 10453}
\ifdefined\arxiv
\author{Hanwen Liang$^{1*}$\quad Shifeng Zhang$^{2}\thanks{Equal contribution. This work was done when the first two authors were interns at Huawei Noah's Ark Lab.}$\quad Jiacheng Sun$^{1}\thanks{Corresponding email: sunjiacheng1@huawei.com.}$\quad Xingqiu He$^1$ \\ \Large \bf Weiran Huang$^1$\quad Kechen Zhuang$^1$\quad Zhenguo Li$^1$\\ \\% All authors must be in the same font size and format. Use \Large and \textbf to achieve this result when breaking a line
    \Large $^1$Huawei Noah's Ark Lab\quad %, lianghanwen1@huawei.com \\
    $^2$TNList, Tsinghua University \\%zhangsf15@mails.tsinghua.edu.cn\\
}
\definecolor{myred}{HTML}{E51400} %red
\definecolor{myblue}{HTML}{0050EF} %cobalt
\definecolor{mygreen}{HTML}{008A00} %emerald
\definecolor{mypurple}{HTML}{AA00FF} %violet
\definecolor{myorange}{HTML}{FF7F00}
\definecolor{mygray}{HTML}{848482}
\usepackage[colorlinks=true,citecolor=myblue,linkcolor=myred,filecolor=mypurple,urlcolor=black]{hyperref}
\fi

\begin{document}
\maketitle
\begin{abstract}
Recently, there has been a growing interest in automating the process of neural architecture design, and the Differentiable Architecture Search (DARTS) method makes the process available within a few GPU days.
% In particular, a hyper-network called one-shot model is introduced, over which the architecture can be searched continuously with gradient descent.
However, the performance of DARTS is often observed to collapse when the number of search epochs becomes large.
Meanwhile, lots of ``{\em skip-connect}s'' are found in the selected architectures.
In this paper, we claim that the cause of the collapse is that there exists overfitting in the optimization of DARTS.
Therefore, we propose a simple and effective algorithm, named ``DARTS+'', to avoid the collapse and improve the original DARTS, by ``early stopping'' the search procedure when meeting a certain criterion.
We also conduct comprehensive experiments on benchmark datasets and different search spaces and show the effectiveness of our DARTS+ algorithm, and DARTS+ achieves $2.32\%$ test error on CIFAR10, $14.87\%$ on CIFAR100, and $23.7\%$ on ImageNet.
We further remark that the idea of ``early stopping'' is implicitly included in some existing DARTS variants by manually setting a small number of search epochs,
while we give an {\em explicit} criterion for ``early stopping''.
\end{abstract}

\section{Introduction}

Neural Architecture Search (NAS) plays an important role in Automated Machine Learning (AutoML), which has attracted a lot of attention recently~\cite{regular2016neural,pham2018efficient,liu2018darts,chen2019progressive,luo2018neural,hang2019auto-fpn,gong2019autogan,liu2020autofis}. Differentiable Architecture Search (DARTS)~\cite{liu2018darts} receives broad attention as it can perform searching very fast while achieves the desired performance. In particular,
it encodes the architecture search space with continuous parameters to form a one-shot model and performs searching by training the one-shot model with gradient-based bi-level optimization.

\begin{figure*}[t]
    % % \setlength{\abovecaptionskip}{0.5pt}
    % % \setlength{\belowcaptionskip}{0pt}
    \begin{center}
        \includegraphics[width=0.9\linewidth]{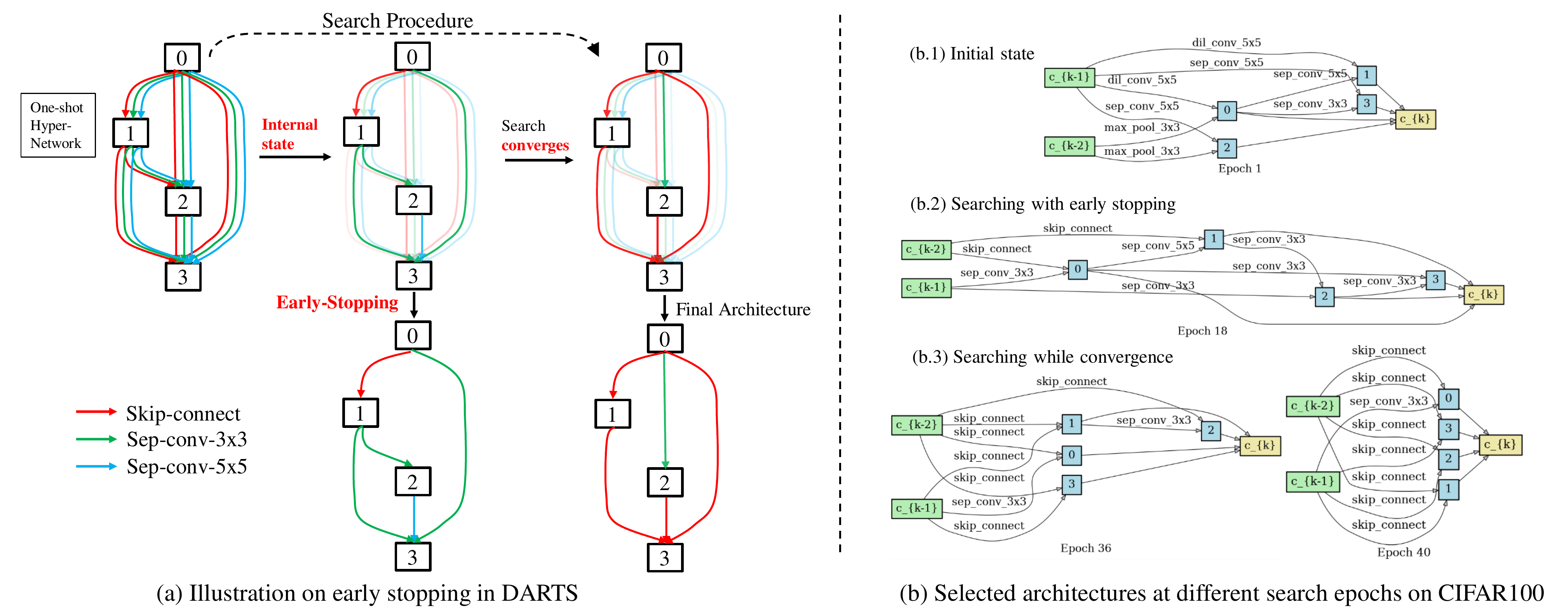}
        \caption{(a) Illustration of the early stopping paradigm. (b) Selected architectures at different search epochs on CIFAR100. With different stopping mechanisms, we may obtain better architectures or end up with collapsed architectures with lots of skip-connects.}
        \label{visulization}
    \end{center}
\end{figure*}

Despite the efficiency of DARTS, a critical issue of DARTS has been found~\cite{chen2019progressive,yang2019evaluation,chu2019fair,zhou2020theory}. Namely, after certain search epochs, the number of {\em skip-connect}s increases dramatically in the selected architecture, which results in poor performance. We call the phenomenon of performance drop after a certain number of epochs the ``collapse'' of DARTS.
To tackle such an issue, some works like P-DARTS~\cite{chen2019progressive} design search space regularization to alleviate the dominance of {\em skip-connect}s during the search.
However, these approaches involve more hyper-parameters, which need to be carefully tuned by human experts.
Moreover, Single-Path NAS~\cite{stamoulis2019single}, StacNAS~\cite{li2019stacnas}, and SNAS~\cite{xie2018snas} use the one-level optimization instead of the bi-level optimization in DARTS, where the architecture parameters and model weights are updated simultaneously.
However, the search spaces of these algorithms need to be carefully designed~\cite{liu2018darts,li2019stacnas,xie2018snas}.
In summary, the mechanism of the collapse of DARTS remains open.

In this paper, we first show that the collapse of DARTS is due to the {\em overfitting} in the search phase, which results in a large gap between training and validation error.
In particular, we explain why {\em overfitting} results in a large number of {\em skip-connect}s in the selected architectures in DARTS, which hurts the performance of the selected architecture.
To avoid the collapse of DARTS, we add a simple and effective ``early stopping'' paradigm, called ``DARTS+'', where the search procedure stops by a certain criterion, illustrated in Fig.~\ref{visulization}.(a). 
We point out that the searching is saturated when it is ``early stopped''.
We remark that some progress of DARTS, including P-DARTS~\cite{chen2019progressive}, Auto-DeepLab~\cite{liu2019auto}, and PC-DARTS~\cite{xu2019pc}, also adopt the early stopping idea implicitly where fewer search epochs are manually set in their methods. 

Moreover, we conduct sufficient experiments to demonstrate the effectiveness of the proposed DARTS+ algorithm.
Specifically, DARTS+ succeeds in searching on various spaces including DARTS, MobileNetV2, ResNet. In the DARTS search space, DARTS+ achieves $2.32\%$ test error on CIFAR10 and $14.87\%$ test error on CIFAR100, while the search time is less than $0.4$ GPU days.
When transferring to ImageNet, DARTS+ achieves $23.7\%$ top-1 error and impressive $22.5\%$ top-1 error if SE-Module~\cite{hu2018squeeze} is introduced.
DARTS+ is also able to search on ImageNet directly
and gets $23.9\%$ top-1 error.

In summary, our main contributions are as follows:
\begin{itemize}
    \item We study the collapse issue of DARTS, and point out the underlying reason is the overfitting of the model weights in the DARTS training.
    \item We introduce an efficient ``early stopping'' paradigm for DARTS to avoid the collapse and propose effective and adaptive criteria for early stopping.
    \item We conduct extensive experiments on benchmark datasets and various search spaces to demonstrate the effectiveness of the proposed algorithm, which achieves state-of-the-art results on all of them.
\end{itemize}

\section{Collapse of DARTS}

There is an undesired behavior of DARTS~\cite{liu2018darts} that too many {\em skip-connect}s tend to appear in the selected architecture when the number of search epochs is large, making the performance poor.
The phenomenon of performance drop is called the ``collapse'' of DARTS in our paper.
In this section, we first give a quick review of the original DARTS, and then point out the collapse issue of DARTS and discuss its underlying causes.

\subsection{DARTS}

%\label{sec:darts}
The goal of DARTS is to search for a cell, which can be stacked to form a convolutional network or a recurrent network.
Each cell is a directed acyclic graph (DAG) of $N$ nodes $\{x_i\}_{i=0}^{N-1}$, where each node represents a network layer.
We denote the operation space as $\mathcal{O}$, and each element is a candidate operation, e.g., {\em zero, skip-connect, convolution, max-pool}, etc.
Each edge $(i,j)$ of DAG represents the information flow from node $x_i$ to $x_j$, which consists of the candidate operations weighted by the architecture parameter $\alpha^{(i,j)}$.
In particular, each edge $(i,j)$ can be formulated by a function $\bar{o}^{(i,j)}$ where
%\begin{equation*}
$\bar{o}^{(i,j)} (x_i) = \sum_{o \in \mathcal{O}} p_o^{(i,j)} \cdot o(x_i),$
%\end{equation*}
and the weight of each operation $o\in\mathcal{O}$ is a softmax of the architecture parameter $\alpha^{(i,j)}$, that is
$p_o^{(i,j)} = \frac{\exp (\alpha_o^{(i,j)})}{\sum_{o' \in \mathcal{O}} \exp (\alpha_{o'}^{(i,j)})}$.
An intermediate node is $x_j=\sum_{i<j} \bar{o}^{(i,j)} (x_i)$, and
the output node $x_{N-1}$ is depth-wise concatenation of all the intermediate nodes excluding input nodes.
%where $\alpha = \{\alpha_{o}^{(i,j)}\}$ is the architecture parameter to learn.
The above hyper-network is called the one-shot model, and we denote $w$ as the weights of the hyper-network.

For the search procedure, we denote $\mathcal{L}_{train}$ and $\mathcal{L}_{val}$ as the training and validation loss respectively.
Then the architecture parameters are learned with the following bi-level optimization problem:
\begin{equation*}
    \begin{split}
\min_\alpha\quad &\mathcal{L}_{val} (w^*(\alpha), \alpha), \\
\mathrm{s.t.}\quad &w^*(\alpha) = \arg \min_w \mathcal{L}_{train} (w, \alpha).
    \end{split}
\end{equation*}

After obtaining architecture parameters $\alpha$, the final discrete architecture is derived by:
1) setting $o^{(i,j)} = \arg \max_{o \in \mathcal{O}, o\neq zero} p_o^{(i,j)}$, and
2) for each intermediate node, choosing two incoming edges with the two largest values of $\max_{o \in \mathcal{O}, o\neq zero} p_o^{(i,j)}$.
More technical details can be found in the original DARTS paper~\cite{liu2018darts}.

\begin{figure*}[t]
    \subfigure[Detailed illustration on the collapse of DARTS. (a) The performance of architectures at different epochs on CIFAR10, CIFAR100, and Tiny-ImageNet-200, respectively (in blue line), and the number of {\em skip-connect}s in the normal cell (in green line). (b) The change of $\alpha$ in the deepest edge (connecting the last two nodes) of the one-shot model. We omit the $\alpha$ of {\em none} operation as it increases while $\alpha$ of other operations drops. (c) The change of architecture parameters $\alpha$ in the shallowest edge. The dashed line denotes the early-stopping paradigm introduced in Sec. \ref{sec:early_stopping}, and the circle denotes the point that the $\alpha$ ranking of learnable parameters becomes stable.]{
        \includegraphics[width=0.63\linewidth]{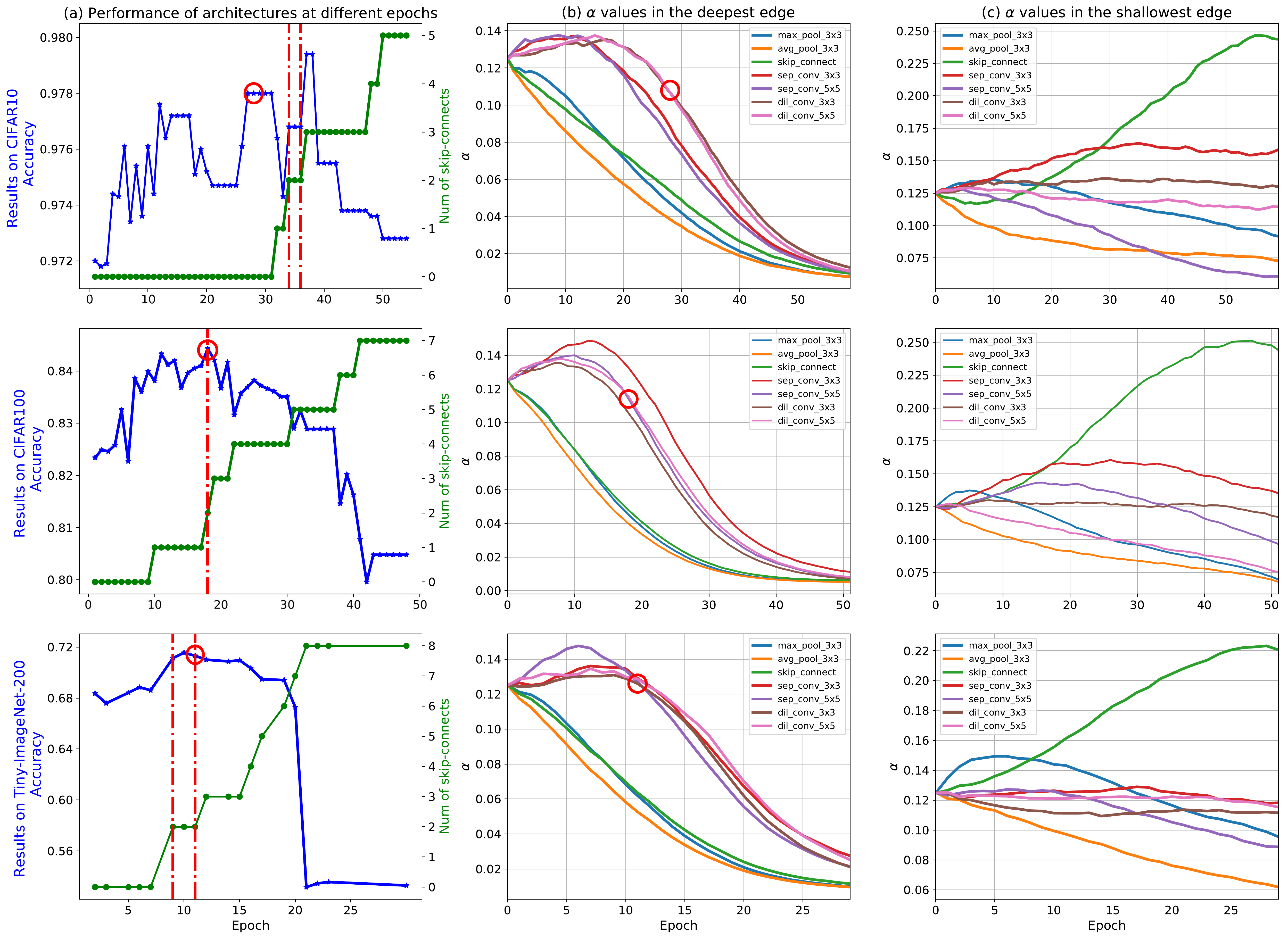}
        \label{all_exp}
    }
    \quad
    \subfigure[The selected architecture of normal cells at different layers when searching distinct cell architectures in different stages (stages are split with reduction cells). The searched dataset is CIFAR100. The first cells contain mostly convolutions, while the last cells are shallow with numerous {\em skip-connect}s.]{
        \includegraphics[width=0.34\linewidth]{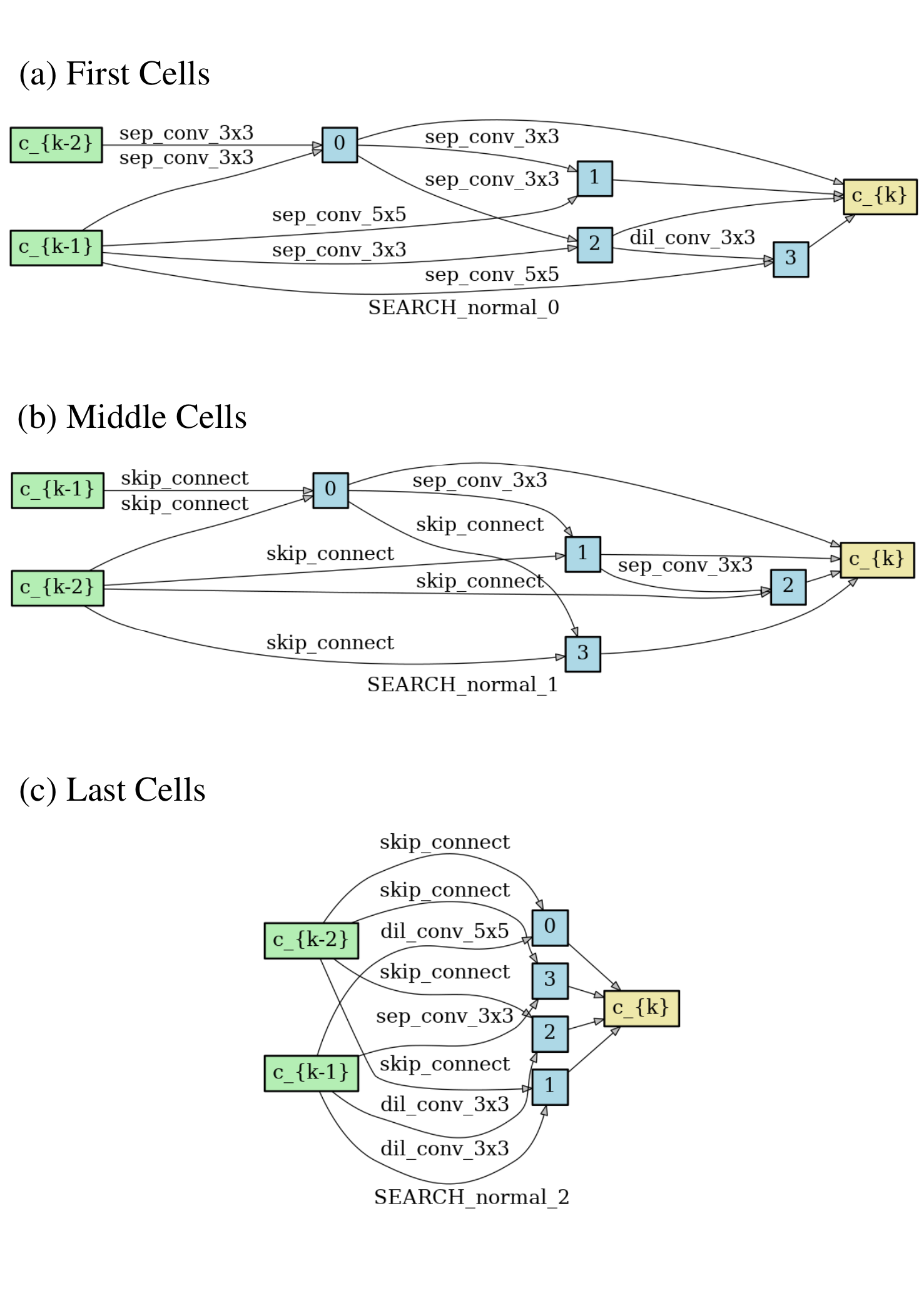}
        \label{fig:diffcell}
    }
    \caption{The collapse issue of DARTS.}
\end{figure*}

\subsection{Collapse Issue}
\label{sec:collapse}

It has been observed in DARTS that lots of {\em skip-connect}s are involved in the selected architecture, which makes the architecture shallow and the performance poor.
As an example, let us consider searching on CIFAR100.
The $\alpha$ value of {\em skip-connect}s (green line in Fig.~\ref{all_exp}(c)) becomes large when the number of search epochs is large,
thus the number of {\em skip-connect}s increases in the selected architecture as shown in the green line in Fig.~\ref{all_exp}(a).
%With lots of {\em skip-connect}s, the selected architecture tends to be a shallow one.
Such a shallow network has less learnable parameters than deep ones, and thus it has weaker expressive power.
As a result, architectures with lots of {\em skip-connect}s have poor performance, i.e., {\em collapsed}, indicated as the blue line in Fig.~\ref{all_exp}(a)\footnote{All the architecture are fully trained and evaluated in the same settings.}, see more experiments in Appendix C.
To be more intuitive, we draw the selected architectures from different search epochs on CIFAR100 in Fig.~\ref{visulization}(b).
When the number of search epochs increases, the number of {\em skip-connect}s in the selected architecture also increases. 
Such a phenomenon can also be observed on other datasets, such as CIFAR10 and Tiny-ImageNet-200.

To avoid the collapse, one might propose to adjust searching hyper-parameters, such as 1) adjusting learning rates, 2) changing the portion of training and validation data, and 3) adding regularization on {\em skip-connect}s like {\em dropout}.
Unfortunately, such methods only alleviate the collapse at certain searching epochs but the collapse would finally appear, implying that the choice of hyperparameters is not the essential cause of the collapse.

\subsection{Overfitting and Analysis}

To figure out the collapse of DARTS, we observe that the model weights in the one-shot model suffer from {\em ``overfitting''} during the search procedure. In the bi-level optimization of DARTS, the model weights $w$ are updated with the training data while the architecture parameters $\alpha$ are updated with the validation data. As the model weights, $w$ in the one-shot model is over-parameterized, $w$ tends to fit the training data well, while the validation data is underfitted as the number of $\alpha$ is limited. To be specific, in CIFAR10/100 dataset, the training accuracy can reach $99\%$ while the validation accuracy is just $88\%$ in CIFAR10 and $60\%$ in CIFAR100. This implies {\em ``overfitting''} as the gap between the training and validation error is large.\footnote{The definition of "overfitting" is slightly different from the general definition such that lower training error results in a higher validation error. However, in both cases, the gap between the training and validation error is large.}

We show that the {\em overfitting} of the model weights is the main cause of the collapse of DARTS. In particular, at the initial state, the model weights underfit the training data and the gap between the training and validation error is small.
Therefore, the architecture parameters $\alpha$ and model weights $w$ get better together. 
% The candidate operations with larger value of $\alpha$ tend do be selected as these operations fit both the training and the validation data.
After certain search epochs, the model weights overfit the training data. However, on validation data, they fit not as well as training data and the first cells of the model could obtain relatively better low-level feature representations than those in the last cells.

If we allow different cells to have distinct architectures in the one-shot model, the last cells are more likely to select more {\em skip-connect}s to obtain the good feature representation directly from the first cells. 
Fig.~\ref{fig:diffcell} shows the learned normal cell architectures at different layers if we search different architectures at different stages\footnote{Stages are split with reduction cells, and each stage consists of several stacked cells.}.
It can be seen that the algorithm tends to select deep architectures with learnable operations in the first cells (Fig.~\ref{fig:diffcell}(a)), while architectures with many {\em skip-connect}s are preferred in the last cells (Fig.~\ref{fig:diffcell}(c)). It accords with the previous analysis such that the last layers will select {\em skip-connect}s.
If different cells are forced to have the same architecture, as DARTS does, continuous searching and fitting will push {\em skip-connect}s to be broadcast from the last cells to the first cells,
making the number of {\em skip-connect}s in the selected architecture grow gradually.
We could see the overfitting of model weights causes the degradation of the selected model architecture. 

\begin{figure}[t]
    \begin{center}
        \includegraphics[width=1.02\linewidth]{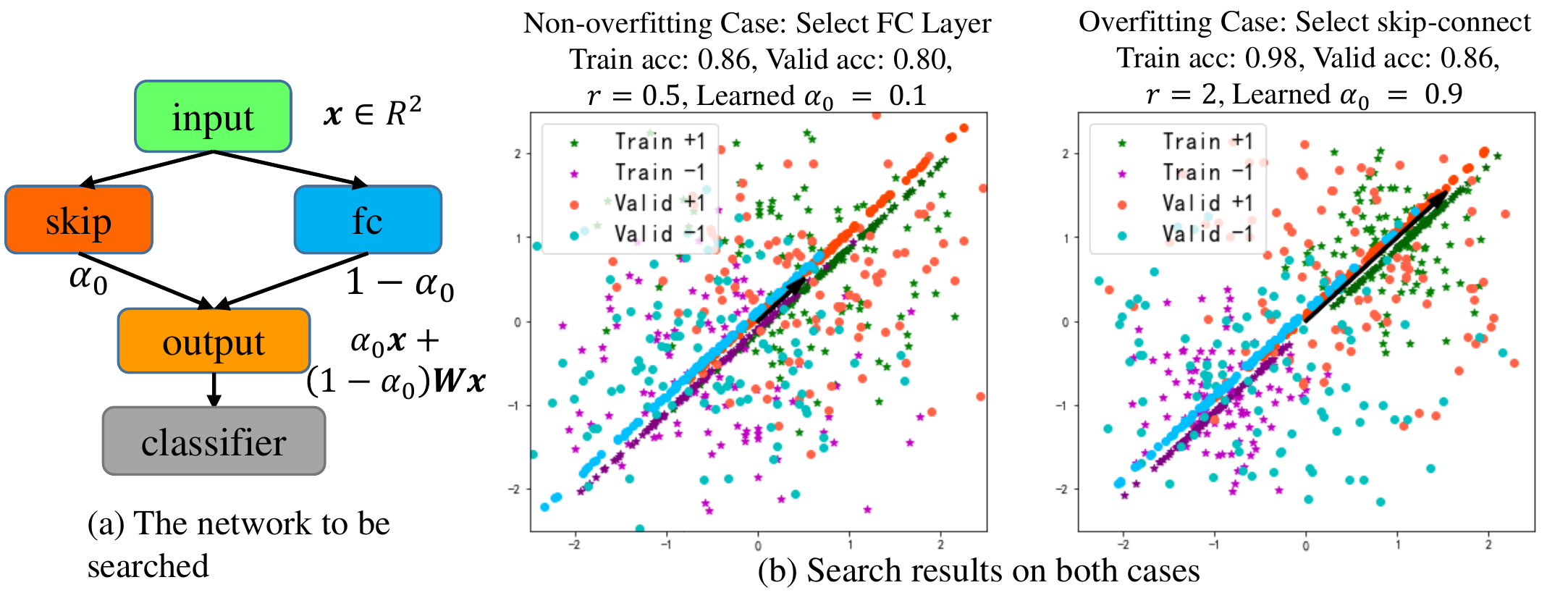}
        \caption{Illustration of overfitting on the synthetic data. (a) The network to be searched, in which the detailed network definition is in Lemma \ref{lemma1}. (b) Search results in both cases. The training/validation input features are a mixture of Gaussians, the features after the fully-connect(fc) layer are along the straight line, and the black arrow denotes $\mathbf{w}_r = r \mathbf{e}$.}
        \label{fig:synthetic}
    \end{center}
\end{figure}

This {\em over-fitting} phenomenon can be further illustrated as a synthetic binary classification problem shown in Fig. \ref{fig:synthetic}. The one-shot model is a 2-layer network defined as $o(\mathbf{x}) = \mathbf{w}_r^\top (\alpha_0 \mathbf{x} + (1-\alpha_0) \mathbf{W} \mathbf{x})$, where $\mathbf{W}, \mathbf{w}_r$ are model weights with $\| \mathbf{w}_r \| = r$, and $\alpha_0$ is the architecture parameter (shown in Fig. \ref{fig:synthetic}(a)). The training data $\mathcal{T}$ and validation data $\mathcal{V}$ used for architecture search are 2-D feature representations and they are mixture of Gaussians such that $\mathcal{T} = \{(\mathbf{x}_i ,y_i), y_i \mathbf{x}_i \sim N(\mu_t \mathbf{e}, \sigma_t^2 \mathbf{I})\}$, $\mathcal{V} = \{(\mathbf{x}_i ,y_i), y_i \mathbf{x}_i \sim N(\mu_v \mathbf{e}, \sigma_v^2 \mathbf{I})\}$ where $\mathbf{e} = \frac{1}{\sqrt{2}}(1,1)^\top$. Both training and validation labels are balanced such that the number of data with label $1$ is the same as that with label $-1$. If the one-shot model is searched with DARTS where the training and validation losses are $\mathcal{L}_{train} = \sum_{(\mathbf{x}_i, y_i) \in \mathcal{T}} l(o(\mathbf{x}_i), y_i), \mathcal{L}_{val} = \sum_{(\mathbf{x}_i, y_i) \in \mathcal{V}} l(o(\mathbf{x}_i), y_i), l(o, y) = \log (1+\exp(-y o))$, then under certain conditions, {\em skip-connect} will be selected, which is summarized in the following lemma.

\begin{lemma}
Consider searching with a binary classification problem where the data and the one-shot model are defined above (shown in Fig. \ref{fig:synthetic}). Suppose (1) the feature representations are normalized such that $\frac{1}{2}\mu_t^2 + \sigma_t^2 =  \frac{1}{2}\mu_v^2 + \sigma_v^2 = 1$
\footnote{Denote $\mathbf{x}[i]$ as the $i$th dimension of $\mathbf{x}$. As the training features are normalized, for the training data, it should be $\mathbb{E}_{\mathcal{T}}[\mathbf{x}] = 0$ and $\mathbb{D}_{\mathcal{T}}[\mathbf{x}[i]] = 1$ for any $i = 0,1$. For both dimension, it is clear that $\mathbb{E}_{\mathcal{T}}[\mathbf{x}] = 0$ as the labels are balanced, and $\mathbb{D}_{\mathcal{T}}[\mathbf{x}[i]] = \frac{1}{2} \mu_t^2 + \sigma_t^2$ for any $i = 0,1$ as $y \mathbf{x}[i] \sim N(\frac{\mu_t}{\sqrt{2}}, \sigma_t^2), (\mathbf{x}, y) \in \mathcal{T}$. Then we have $\frac{1}{2}\mu_t^2 + \sigma_t^2 = 1$. The same holds for the validation data such that $\frac{1}{2}\mu_v^2 + \sigma_v^2 = 1$.}, and the outputs of the fully-connected (fc) layer $\{\mathbf{W} \mathbf{x}\}$ are normalized as above; (2) the above losses are defined and are trained with bi-level optimization. Then we have

\textbf{P1}: If $\sigma_t$ is small, for any $\alpha_0$, $\mathbf{w}_r \to r\mathbf{e}, \mathbf{W} \to \mathbf{W}^*$, where $\mathbf{W}^* \mathbf{x} = t_{\mathbf{x}} \mathbf{e}$, and $ t_{\mathbf{x}} = \frac{2}{\sqrt{2 + \mu_t^2}} \mathbf{e}^\top \mathbf{x}.$

\textbf{P2} If $\sigma_t$ is small and $\sigma_v > \sigma_0 (r) $ where $\sigma_0 (r) $ is a monotonic decreasing function, then $\frac{\mathrm{d} \mathcal{L}_{val}}{\mathrm{d} \alpha_0} < 0$, which implies that $\alpha_0$ will become larger with gradient descent.
\label{lemma1}
\end{lemma}

The proof can be found in Appendix A. In this paper, the feature representations discussed in Lemma \ref{lemma1} correspond to the last layer feature representations in the one-shot model. To be specific, in the beginning of the search procedurem $\sigma_v$ are large and $r$ is small; When overfitting occurs, $r$ goes larger to make the training data more separable while $\sigma_v$ stays large. $\sigma_t$ tend to be small during searching. According to Lemma \ref{lemma1}, in the beginning of searching, $\sigma_v$ is large while $r$ is small and therefore $\sigma_v < \sigma_0(r)$, then learnable operations is preferred (left figure in Fig. \ref{fig:synthetic}(b)). When overfitting occurs, $\sigma_v$ stays relatively large such that $\sigma_v > \sigma_0 (r)$ as larger $r$ makes $\sigma_0 (r)$ small, then {\em skip-connect} tend to be selected (right figure in Fig. \ref{fig:synthetic}(b)).

% Second, the {\em overfitting} phenomenon can also be observed in other bi-level optimization problems like Generative Adversarial Network (GAN)~\cite{goodfellow2014generative}.
% It is proved that a good learned discriminator is essential for training the generator in GAN.
% However, as the input data (fake or real) lies in low-dimensional manifold and the discriminator is over-parameterized, the discriminator is easy to overfit such that it separates the generated fake data from the real, and the generator will suffer from gradient vanishing and fail to generate real data~\cite{arjovsky1701towards}. 

\section{The Early Stopping Methodology}
\label{sec:early_stopping}

Since the collapse issue of DARTS is caused by {\em ``overfitting''} of the one-shot model in the bi-level optimization as pointed out in Sec.~\ref{sec:collapse}, we propose a simple and effective ``early stopping'' paradigm based on DARTS to avoid the collapse.
In particular, the search procedure should be early stopped at an adaptive criterion, when DARTS starts to collapse. Such a paradigm leads to both better performance and fewer search costs than the original DARTS. We use DARTS+ to denote the DARTS algorithm with our early stopping criterion.

We want to emphasize that early stopping is essential and more attention should be paid. It has been found that important connections are determined in the early phase of training~\cite{achille2018critical}. Significant and consequential changes occur during the earliest stage of training~\cite{frankle2019early-phase}. 

Besides the {\em ``overfitting''} issue, another motivation of ``early stopping'' is that the ranking of architecture parameters $\alpha$s of operations matters as only the operation with the maximum $\alpha$ is chosen in the selected architecture. 
%Also, it is pointed out by~\cite{chen2019progressive} that learnable operations enhance the model capability, thus we should give more attention to learnable operations  (e.g., {\em convolutions}). 
During searching, the validation data has different preferences on learnable operations, which corresponds to the ranking of $\alpha$ values. If the rank of $\alpha$ is not stable, the architecture is too noisy to be chosen; while when it becomes stable, the learnable operations in the final selected architecture are not changed, and we could consider this point as the saturated search point. The red circles in Fig.~\ref{all_exp}(a-b) denote the saturated search points on different datasets. It verifies that after this point the validation accuracies of selected architectures on all datasets (blue lines) tend to decrease, i.e., collapse. 
To conclude, the search procedure can be ``early-stopped'' at the saturated searching point to select desired architectures as well as avoiding overfitting, and we emphasize that this point does not mean the convergence of the one-shot model.

We first of all follow the cell-based architecture used by DARTS.
The first criterion is stated as follows.

\vspace{0.2cm}
\noindent \textbf{Criterion 1} {\em The search procedure stops when there are two or more than two {\em skip-connect}s in one normal cell. }
\vspace{0.1cm}

The major advantage of the proposed stopping criterion is its simplicity.
Compared with other DARTS variants, DARTS+ only needs a few modifications based on DARTS, and can significantly increase the performance with less search time.
As too many {\em skip-connect}s will hurt the performance of DARTS, while an appropriate number of {\em skip-connect}s helps to transfer the information from the first layers to the last layers and stabilizing the training process, e.g., ResNet~\cite{he2016deep}, which makes the architectures achieve better performance.
Therefore, stopping by Criterion 1 is a reasonable choice.

The hyper-parameter {\em two} in Criterion 1 is motivated by P-DARTS~\cite{chen2019progressive}, where the number of {\em skip-connect}s in the cell of final architecture is manually cut down to two.
However, DARTS+ is essentially different from P-DARTS in dealing with the {\em skip-connect}s.
\mbox{P-DARTS} does not intervene the number of {\em skip-connect}s during the search procedure, but only replace the redundant {\em skip-connect}s with other operations as post-processing after the search procedure finishes. In contrast, our DARTS+ ends up with desired architectures with a proper number of {\em skip-connect}s to avoid the collapse of DARTS. It controls the number of {\em skip-connect}s more directly and also more effectively (See Table~\ref{tab:cifar_results} for a performance comparison between DARTS+ and P-DARTS).

Since the stable ranking of architecture parameters $\alpha$ for learnable operations indicates the saturated search procedure in DARTS, we can also use the following stopping criterion:

\vspace{0.2cm}
\noindent \textbf{Criterion 2} {\em The search procedure stops when the ranking of architecture parameters $\alpha$ for learnable operations becomes stable for a determined number of epochs (e.g., 10 epochs).}
\vspace{0.1cm}

It can be seen from Fig.~\ref{all_exp} that the saturated training point (stopping point with Criterion 2) is close to the stopping point when Criterion 1 holds (the red dash line in Fig.~\ref{all_exp}(a)).
% thus stopping by Criterion 1 is reasonable as the searching is saturated at this stopping point. 
We also remark that both criteria can be used freely as the stopping points are close. However, Criterion 1 is much easier to operate, but if one needs stopping more precisely or other search spaces are involved, Criterion 2 could be used instead. {\em Ten} epochs in Criterion 2 is a hyper-parameter and according to our experiments, when the ranking of operators remains the same for more than 6 epochs it can be considered stable, implying that this hyper-parameter is not sensitive and flexible to choose. We further remark that our early stopping paradigm solves an intrinsic issue of DARTS and is orthogonal to other tricks, thus it has the potential to be used in other DARTS-based algorithms to achieve better performance. Moreover, our method is very easy to complement than other methods like computing the eigenvalues of Hessian in validation loss~\cite{zela2019understanding}.

\begin{table*}[t]
	\centering
    \tiny
	\caption{Results of different architectures on CIFAR10 and CIFAR100. $^1$~denotes training without cutout augmentation. $^2$~denotes re-implementing with the proposed experimental settings. $^3$~denotes using a different search space from others. $^4$~denotes results of the best architecture searched from the corresponding dataset and training with more channels and more augmentations. * denotes using stopping Criterion 2, otherwise using Criterion 1.}
	\resizebox{\textwidth}{!}{\begin{tabular}{ccccccc}
		\hline
		\multicolumn{1}{c}{\multirow{2}{*}{Architecture}} & Search & \multicolumn{2}{c}{Test Err. (\%)} & Param & Search Cost & Search \\
		\cline{3-4}
		\multicolumn{1}{c}{} & Dataset & CIFAR10 & CIFAR100 & (M) & (GPU days) & \multicolumn{1}{c}{Method} \\
		\hline
		DenseNet-BC~\cite{huang2017densely}$^1$ & - & $3.46$ & $17.18$ & $25.6$ & - & manual \\
		\hline
		NASNet-A~\cite{regular2016neural} & CIFAR10 & $2.65$ & - & $3.3$ & $1800$ & RL \\
		% AmoebaNet-A + cutout & $3.34 \pm 0.06$ & - & $3.2$ & $3150$ & evolution \\
		AmoebaNet-B~\cite{real2019regularized} & CIFAR10 & $2.55 \pm 0.05$ & - & $2.8$ & $3150$ & evolution \\
		PNAS~\cite{liu2018progressive}$^1$ & CIFAR10 & $3.41 \pm 0.09$ & - & $3.2$ & $225$ & SMBO \\
		ENAS~\cite{pham2018efficient} & CIFAR10 & $2.89$ & - & $4.6$ & $0.5$ & RL \\
		NAONet~\cite{luo2018neural} & CIFAR10 & $3.18^{1}$ & $15.67$ & $10.6$ & $200$ & NAO \\
		\hline
		DARTS~\cite{liu2018darts} & CIFAR10 & $3.00$ & $17.76$ & $3.3$ & $1.5$ & gradient \\
		DARTS$^2$ & CIFAR10 & $2.72$ & $16.97$ & $3.3$ & $1.5$ & gradient \\
		SNAS (moderate)~\cite{xie2018snas} & CIFAR10 & $2.85$ & - & $2.8$ & $1.5$ & gradient \\
		ProxylessNAS~\cite{cai2018proxylessnas}$^3$ & CIFAR10 & $2.08$ & - & $5.7$ & $4$ & gradient \\
		P-DARTS~\cite{chen2019progressive} & CIFAR10 & $2.50$ & $16.55$ & $3.4$ & $0.3$ & gradient \\
		P-DARTS~\cite{chen2019progressive} & CIFAR100 & $2.62$ & $15.92$ & $3.6$ & $0.3$ & gradient \\
		ASAP~\cite{noy2019asap} & CIFAR10 & $2.49 \pm 0.04$ & $15.6$ & $2.5$ & $0.2$ & gradient \\
		PC-DARTS~\cite{xu2019pc} & CIFAR10 & $2.57 \pm 0.07$ & - & $3.6$ & $0.1$ & gradient \\
		\hline
		\textbf{DARTS+} & CIFAR10 & $\mathbf{2.32} (2.50 \pm 0.11)$ & $16.28$ & $3.7$ & $0.4$ & gradient \\
		\textbf{DARTS+} & CIFAR100 & $2.46$ & $\mathbf{14.87} (15.42 \pm 0.30)$ & $3.8$ & $0.2$ & gradient \\
		\textbf{DARTS+*} & CIFAR10 & $\mathbf{2.20} (2.37 \pm 0.13)$ & $15.04$ & $4.3$ & $0.6$ & gradient \\
		\textbf{DARTS+*} & CIFAR100 & $2.46$ & $\mathbf{14.87} (15.45 \pm 0.30)$ & $3.9$ & $0.5$ & gradient \\
		\hline
		\textbf{DARTS+ (Large)}{$^4$} & - & $\mathbf{1.68}$ & $\mathbf{13.03}$ & $7.2$ & - & gradient \\
		\hline
	\end{tabular}}
	\label{tab:cifar_results}
\end{table*}

We note that recent state-of-the-art differentiable architecture search methods also introduce the early stopping idea in an ad hoc manner.
To avoid the collapse, P-DARTS~\cite{chen2019progressive} uses 1) searching for 25 epochs instead of 50 epochs, 2) adopting {\em dropout} after {\em skip-connect}s, and 3) manually reducing the number of {\em skip-connect}s to two.
Auto-DeepLab~\cite{liu2019auto} uses fewer epochs for searching the architecture parameters and finds that searching for more epochs does not bring benefits.
PC-DARTS~\cite{xu2019pc} uses partial-channel connections to reduce search time, and therefore more epochs are needed for convergence of searching. Thus, setting 50 training epochs is also an implicit early stopping paradigm, see Appendix C.

\section{Experiments and Analysis}\label{Exp_set}

\subsection{Datasets}

In this section, we conduct extensive experiments on benchmark classification datasets to evaluate the effectiveness of the proposed DARTS+ algorithm. We use four popular datasets including CIFAR10~\cite{krizhevsky2009cifar}, CIFAR100~\cite{krizhevsky2009cifar}, Tiny-ImageNet-200\footnote{\url{https://tiny-ImageNet.herokuapp.com/}} and ImageNet~\cite{deng2009ImageNet}. CIFAR10/100 consists of 50K training images and 10K testing images and the resolution is $32 \times 32$. Tiny-ImageNet-200 contains 100K $64 \times 64$ training images and 10K testing images. ImageNet is obtained from ILSVRC2012~\cite{russakovsky2015ImageNet}, which contains more than 1.2M training images and 50K validation images. We follow the general setting on the ImageNet dataset where the images are resized to $224 \times 224$ for training and testing.

\subsection{Effectiveness of Early Stopping on Different Search Spaces}
\label{sec:diffstop}

To verify the effectiveness of early stopping in DARTS+, we conduct extensive experiments with different datasets on selected architectures at different epochs. The experiments are carried out in two stages: architecture search and architecture evaluation. 

\begin{figure}
    \centering
    \includegraphics[width=0.97\linewidth]{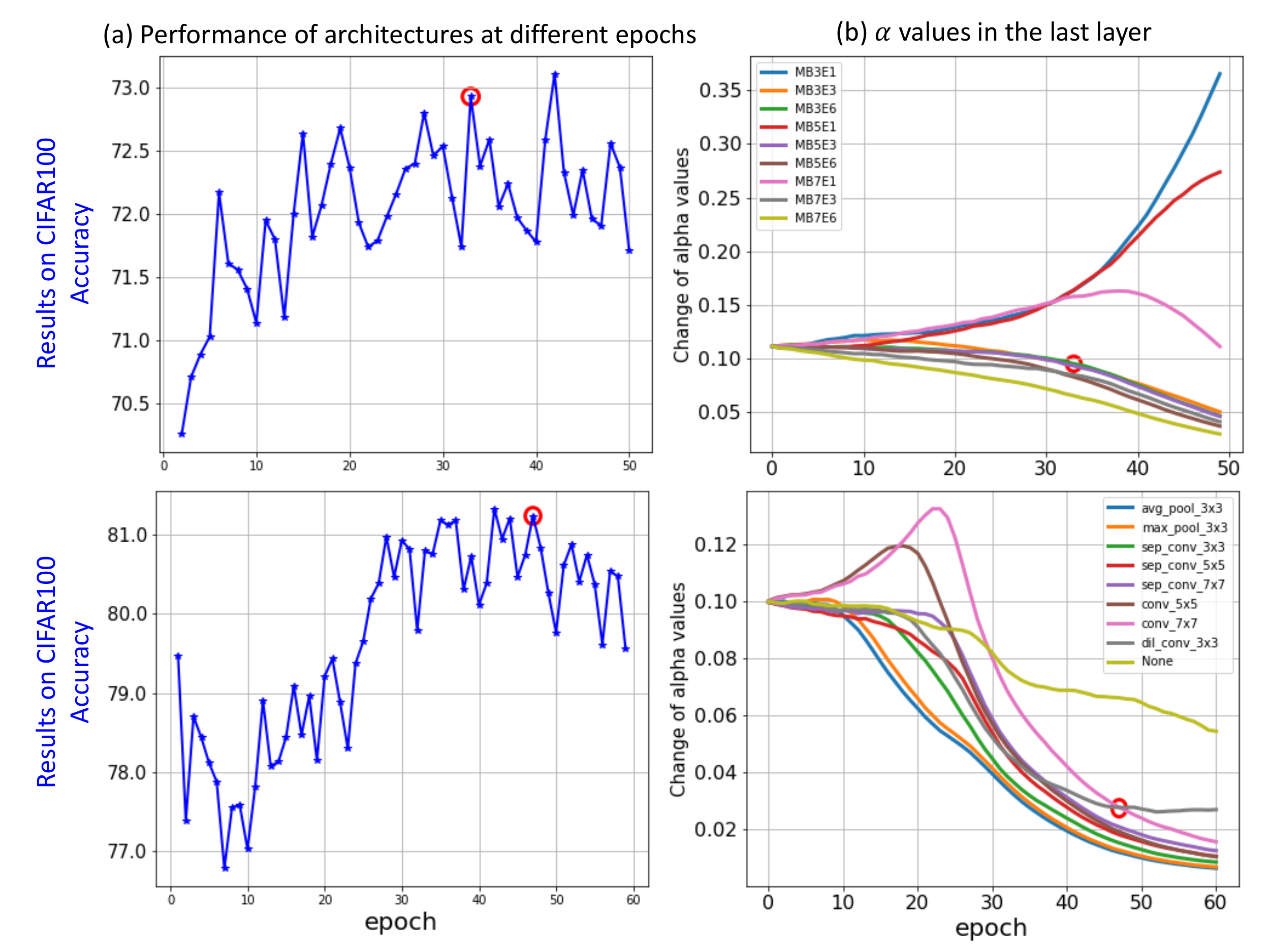}
    \caption{Results on the MobileNetV2 and ResNet search space. (a)(b) denote roughly the same as that in Figure \ref{all_exp}. Red circle denotes the early stopping point at Criterion 2.}
    \label{fig:mbv2_resnet}
\end{figure}

\subsubsection{DARTS Search Space.}

In the DARTS search space, the experimental settings are similar to DARTS. For CIFAR10 and CIFAR100, in the architecture search phase, we use the same one-shot model as the original DARTS and the hyperparameters are almost the same as DARTS except that a maximum of $60$ epochs is adopted. In the architecture evaluation phase, the experimental settings follow the original DARTS, except that 2000 epochs are used for better convergence. For Tiny-ImageNet-200, in the search phase, the one-shot model is almost the same as CIFAR10/100 except that a $3\times3$ convolution layer with stride 2 is added on the first layer to reduce the input resolution from $64 \times 64$ to $32 \times 32$. Other settings are the same as those used in CIFAR10/100. 

We use Criterion 1 and 2 for early stopping in the DARTS search space. Other details of the experimental settings can be referred to in Appendix B.

The classification results of the selected architectures at different epochs are shown in Fig.~\ref{all_exp}. We also mark the ``early stop'' point under two criteria as ``red dashed line'' and ``red circle'' respectively. We observe that the selected architecture performs worse with larger epochs, implying that the original DARTS suffers from the collapse issue. In contrast, ``early stopping'' can generate good architectures at both stopping criteria, regardless of the type of datasets.

We also compare ``early stopping'' Criterion 1 and 2 in Table \ref{tab:cifar_results} and Fig. \ref{all_exp}. We observe that both criteria achieve comparable performance on all datasets as the stopping points are very close.

\begin{table}[t]
	\small
	\caption{Results of different architectures on Tiny-ImageNet-200. $^\dag$~denotes directly searching on Tiny-ImageNet-200, otherwise transferred from CIFAR10. * denotes using Criterion 2, otherwise using Criterion 1}
	\centering
	\begin{tabular}{ccc}
		\hline
		Architecture & Test Err. (\%) & Params (M) \\
		\hline
		ResNet-18~\cite{yao2019differentiable} & $47.3$ & $11.7$ \\
		DenseNet-BC~\cite{lan2018self} & $37.1$ & - \\
		\hline
		NASNet & $29.8$ & $4.5$ \\
		DARTS & $30.4$ & $3.8$ \\
		DARTS$^\dag$ & $46.1$ & $2.1$ \\
		SNAS & $30.6$ & $3.4$ \\
		ASAP & $30.0$ & $3.3$ \\
		\hline
		\textbf{DARTS+} & $\mathbf{29.1}$ & $4.2$ \\
		\textbf{DARTS+}$^\dag$ & $\mathbf{28.3}$ & $3.8$ \\
	    \textbf{DARTS+*}$^\dag$ & $\mathbf{27.6}$ & $4.3$ \\
	\hline
	\end{tabular}
	\label{tab:tiny_imagenet_resutls}
\end{table}

\begin{table*}[t]
	\centering
	\small
	\caption{Results of different architectures on ImageNet. $^*$~denotes re-implementing the result. $^\dag$~denotes directly searching on ImageNet. $^\ddag$~denotes using SE-Module and training with more augmentations (AutoAugment, mixup, etc.)}
	\begin{tabular}{ccccccc}
		\hline
		\multirow{2}{*}{Architecture} & \multicolumn{2}{c}{Test Err. (\%)} & Params  & $\times +$  & Search Cost & \multirow{2}{*}{Search Method} \\
		\cline{2-3}
		& Top-1 & Top-5 & (M) & (M) & (GPU days)  &  \\
		\hline
		% Inception-V1 & $30.2$ & $10.1$ & $6.6$ & $1448$ & - & manual \\
		MobileNet~\cite{howard2017mobilenets} & $29.4$ & $10.5$ & $4.2$ & $569$ & - & manual \\
		MobileNet-V2 ($1.4\times$)~\cite{sandler2018mobilenetv2} & $25.3$ & - & $6.9$ & $585$ & - & manual \\
		ShuffleNet-V2 ($2\times$)~\cite{ma2018shufflenet} & $25.1$ & - & $7.4$ & $591$ & - & manual \\
		\hline
		NASNet-A~\cite{regular2016neural} & $26.0$ & $8.4$ & $5.3$ & $564$ & $1800$ & RL \\
		AmoebaNet-C~\cite{real2019regularized} & $24.3$ & $7.6$ & $6.4$ & $570$ & $3150$ & RL \\
		PNAS~\cite{liu2018progressive} & $25.8$ & $8.1$ & $5.1$ & $588$ & $225$ & SMBO \\
		MnasNet-92~\cite{tan2019mnasnet} & $25.2$ & $8.0$ & $4.4$ & $388$ & - & RL \\
		EfficientNet-B0~\cite{tan2019efficientnet} & $23.7$ & $6.8$ & $5.3$ & $390$ & - & RL \\
		\hline
		DARTS~\cite{liu2018darts} & $26.7$ & $8.7$ & $4.7$ & $574$ & $4.0$ & gradient \\
		SNAS (mild)~\cite{xie2018snas} & $27.3$ & $9.2$ & $4.3$ & $522$ & $1.5$ & gradient \\
		ProxylessNAS~\cite{cai2018proxylessnas}$^\dag$ & $24.9$ & $7.5$ & $7.1$ & $465$ & $8.3$ & gradient \\
		P-DARTS (CIFAR10)~\cite{chen2019progressive} & $24.4$ & $7.4$ & $4.9$ & $557$ & $0.3$ & gradient \\
		% BayesNAS & $26.5$ & $8.9$ & $3.9$ & - & $0.2$ & gradient \\
		ASAP~\cite{noy2019asap} & $26.7$ & - & - & - & $0.2$ & gradient \\
		XNAS~\cite{nayman2019xnas} & $24.0$ & - & $5.2$ & $600$ & $0.3$ & gradient \\
		PC-DARTS~\cite{xu2019pc}$^\dag$ & $24.2$ & $7.3$ & $5.3$ & $597$ & $3.8$ & gradient \\
		PC-DARTS$^{*\dag}$ & $23.8$ & $7.3$ & $5.3$ & $597$ & $3.8$ & gradient \\
		\hline
		\textbf{DARTS+ (CIFAR100)} & $\mathbf{23.7}$ & $\mathbf{7.2}$ & $5.1$ & $591$ & $0.2$ & gradient \\
		\textbf{DARTS+$^{\dag}$} & $23.9$ & $7.4$ & $5.1$ & $582$ & $6.8$ & gradient \\
		% DARTS+ (CIFAR100)$^\ddag$ & $23.6$ & $6.9$ & $5.1$ & $591$ & $0.2$ & gradient \\
		\textbf{SE-DARTS+ (CIFAR100)}$^\ddag$ & $\mathbf{22.5}$ & $\mathbf{6.4}$ & $6.1$ & $594$ & $0.2$ & gradient \\
		\hline
	\end{tabular}
	\label{tab:imagenet_results}
\end{table*}

\subsubsection{MobileNetV2 and ResNet Search Space.}

To further verify the effectiveness of DARTS+, we use MobileNetV2~\cite{sandler2018mobilenetv2} and ResNet~\cite{he2016deep} as the backbone to build the architecture space~\cite{cai2018proxylessnas}. For the MobileNetV2 search space, We introduce a set of mobile inverted bottleneck convolution (MBConv) with various kernel sizes and expansion ratios to construct the searching block. For ResNet search space, we construct the one-shot model by replacing the residual block with a set of candidate operations, where we keep the skip-connect in the residual block and involve 10 candidate operations. In both search spaces, softmax is applied to architecture parameters to compute the weights, which are used to determine the selected architecture. The experiments are conducted on CIFAR100 dataset. Details of the search spaces and experimental settings on architecture search and architecture evaluation are summarized in Appendix B.

As the {\em skip-connect}s are not involved in the search spaces, we use ``early stopping'' Criterion 2. The classification results of the selected architectures at different epochs are shown in Fig.~\ref{fig:mbv2_resnet}. The time to ``early stop'' with Criterion 2 is marked as ``red circle''. It can be seen that the selected architecture with ``early stopping'' achieves relatively the best performance, compared with randomly searched architecture (epoch 0) and that in large epochs.

\subsection{Comparison with State-of-the-Art}
\label{sec:search_exp}

Unless specified, we use the DARTS search space and ``early stopping'' Criterion 1 evaluating DARTS+.
Note that the stopping points by Criterion~1 and 2 are almost the same in the proposed search space, as discussed in Sec.~\ref{sec:diffstop}.

For CIFAR10, CIFAR100, and Tiny-Imagenet-200 datasets, the experimental settings on both architecture search and architecture evaluation phase can be found in Sec. \ref{sec:diffstop} and Appendix B. For ImageNet, following~\cite{xu2019pc}, the one-shot model starts with three $3\times 3$ convolution layers with stride 2 to reduce the resolution from $224 \times 224$ to $28 \times 28$, and the rest of the network consists of $8$ cells. We select $10\%$ data from the training set for updating model weights, and another $10\%$ for updating architecture parameters. In the architecture evaluation phase, we train the model for 800 epochs with batch size 2048 for better convergence. Other experimental settings are almost the same as those in DARTS, which can be found in Appendix B.

\subsubsection{Search Results and Analysis.}

The proposed DARTS+ needs less searching time as ``early stopping'' is adopted. For CIFAR10, the search procedure requires 0.4 GPU days with a single Tesla V100 GPU and stops at about epoch 35. For CIFAR100, the searching time is 0.2 GPU days and the search procedure stops at about epoch 18. For Tiny-ImageNet-200, searching stops at about epoch 10. For ImageNet, the search procedure involves 200 epochs and requires 6.8 GPU days on Tesla P100 GPU. 

The number of {\em skip-connect}s shown in Fig. \ref{all_exp} implies that the cells searched by DARTS+ contain a small number of {\em skip-connect}s, showing that DARTS+ succeeds in searching with all three datasets including CIFAR10/100, Tiny-ImageNet-200, and ImageNet. However, the original DARTS fails to search on CIFAR100 as the selected architecture is full of {\em skip-connect}s, and most previous works on differentiable search~\cite{liu2018darts,xie2018snas,chen2019progressive} do not search on ImageNet.

The selected architecture can be found in Appendix B.

\subsubsection{Architecture Evaluation on CIFAR10 and CIFAR100.}

The evaluation results are summarized in Table \ref{tab:cifar_results}. For each selected cell from either CIFAR10 or CIFAR100, we report the performance on both datasets. With the simple ``early stopping'' paradigm, we achieve the best results with $2.32\%$ test error on CIFAR10 and $14.87\%$ test error on CIFAR100. The proposed DARTS+ is much simpler and better than other modified DARTS algorithms like P-DARTS and PC-DARTS. ProxylessNAS uses a different search space, and it involves more search time. Moreover, DARTS+ is much easier to implement than other modified DARTS variants including ASAP.

We further increase the initial channel number from 36 to 50 and add more augmentation tricks including AutoAugment~\cite{cubuk2018autoaugment} and mixup~\cite{zhang2017mixup} to achieve better results. Table \ref{tab:cifar_results} shows that DARTS+ achieves impressive $1.68\%$ test error on CIFAR10 and $13.03\%$ on CIFAR100, demonstrating the effectiveness of DARTS+.

\subsubsection{Architecture Evaluation on Tiny-ImageNet-200.}

For DARTS and DARTS+, we use the architecture searched directly from Tiny-ImageNet-200 for evaluation. We also transfer the architectures searched from other algorithms for a fair comparison. The results are shown in Table \ref{tab:tiny_imagenet_resutls}. DARTS+ achieves the state-of-the-art $28.3\%$ test error with Criterion 1 and $27.6\%$ test error with Criterion 2. Note that architecture searched on Tiny-ImageNet-200 with DARTS has less parameter size and performs much inferior, because DARTS suffers from collapse and the architecture searched with DARTS contains lots of {\em skip-connect}s.

\subsubsection{Architecture Evaluation on ImageNet.}

We use the architecture searched directly from ImageNet for evaluation, and the architecture from CIFAR100 to test the transferability of the selected architecture. 
The experimental results are shown in Table \ref{tab:imagenet_results}. Note that we re-implement PC-DARTS and the results are reported. When searching on ImageNet with the proposed DARTS+, the selected architecture achieves an impressive $23.9\%/7.4\%$ top-1/top-5 error, and the architecture transferred from CIFAR100 achieves state-of-the-art $23.7\%/7.2\%$ error. The results imply that DARTS with ``early stopping'' succeeds in searching for a good architecture with an impressive performance on large-scale datasets with limited time.

We also adopt SE-module~\cite{hu2018squeeze} in the architecture transferred from CIFAR100, and introduce AutoAugment~\cite{cubuk2018autoaugment} and mixup~\cite{zhang2017mixup} for training to obtain a better model. The results are shown in Table \ref{tab:imagenet_results}, and we achieve $22.5\%/6.4\%$ top-1/top-5 error with only additional 3M flops, showing the effectiveness of the selected architecture.

\section{Conclusion}

In this paper, we conduct a comprehensive analysis and extensive experiments to show that DARTS suffers from the collapse problem, which is mainly caused by the {\em overfitting} of the one-shot model in DARTS. We propose ``DARTS+'', in which the ``early stopping'' paradigm is introduced to avoid the collapse of DARTS. The experiments show that we succeed in searching on various benchmark datasets including large-scale ImageNet with limited GPU days, and the resulting architectures achieve state-of-the-art performance on all benchmark datasets. Moreover, the proposed ``early stopping'' criteria could be applied to 
various search spaces and many recent signs of progress of DARTS could use ``early stopping'' to achieve better results.
    % \clearpage
    
    %\subsubsection*{Acknowledgments}
    %%Use unnumbered third level headings for the acknowledgments. All acknowledgments
    %go at the end of the paper. Do not include acknowledgments in the anonymized
    %submission, only in the final paper.

    %\medskip

\bibliographystyle{aaai}
\bibliography{mybibfile}

\begin{thebibliography}{}

\bibitem[\protect\citeauthoryear{Achille, Rovere, and
  Soatto}{2019}]{achille2018critical}
Achille, A.; Rovere, M.; and Soatto, S.
\newblock 2019.
\newblock Critical learning periods in deep networks.
\newblock {\em ICLR 2019}.

\bibitem[\protect\citeauthoryear{Cai, Zhu, and Han}{2019}]{cai2018proxylessnas}
Cai, H.; Zhu, L.; and Han, S.
\newblock 2019.
\newblock Proxylessnas: Direct neural architecture search on target task and
  hardware.
\newblock In {\em 7th International Conference on Learning Representations,
  {ICLR}}.

\bibitem[\protect\citeauthoryear{Chen \bgroup et al\mbox.\egroup
  }{2019}]{chen2019progressive}
Chen, X.; Xie, L.; Wu, J.; and Tian, Q.
\newblock 2019.
\newblock Progressive differentiable architecture search: Bridging the depth
  gap between search and evaluation.
\newblock In {\em Proceedings of the IEEE International Conference on Computer
  Vision},  1294--1303.

\bibitem[\protect\citeauthoryear{Chu \bgroup et al\mbox.\egroup
  }{2019}]{chu2019fair}
Chu, X.; Zhou, T.; Zhang, B.; and Li, J.
\newblock 2019.
\newblock Fair darts: Eliminating unfair advantages in differentiable
  architecture search.
\newblock {\em arXiv preprint arXiv:1911.12126}.

\bibitem[\protect\citeauthoryear{Cubuk \bgroup et al\mbox.\egroup
  }{2018}]{cubuk2018autoaugment}
Cubuk, E.~D.; Zoph, B.; Mane, D.; Vasudevan, V.; and Le, Q.~V.
\newblock 2018.
\newblock Autoaugment: Learning augmentation policies from data.
\newblock {\em arXiv preprint arXiv:1805.09501}.

\bibitem[\protect\citeauthoryear{Deng \bgroup et al\mbox.\egroup
  }{2009}]{deng2009ImageNet}
Deng, J.; Dong, W.; Socher, R.; Li, L.-J.; Li, K.; and Fei-Fei, L.
\newblock 2009.
\newblock Imagenet: A large-scale hierarchical image database.
\newblock In {\em CVPR}.

\bibitem[\protect\citeauthoryear{Frankle, Schwab, and
  Morcos}{2020}]{frankle2019early-phase}
Frankle, J.; Schwab, D.~J.; and Morcos, A.~S.
\newblock 2020.
\newblock The early phase of neural network training.
\newblock In {\em ICLR}.

\bibitem[\protect\citeauthoryear{Gong \bgroup et al\mbox.\egroup
  }{2019}]{gong2019autogan}
Gong, X.; Chang, S.; Jiang, Y.; and Wang, Z.
\newblock 2019.
\newblock Autogan: Neural architecture search for generative adversarial
  networks.
\newblock In {\em Proceedings of the IEEE International Conference on Computer
  Vision},  3224--3234.

\bibitem[\protect\citeauthoryear{Goyal \bgroup et al\mbox.\egroup
  }{2017}]{goyal2017accurate}
Goyal, P.; Doll{\'a}r, P.; Girshick, R.; Noordhuis, P.; Wesolowski, L.; Kyrola,
  A.; Tulloch, A.; Jia, Y.; and He, K.
\newblock 2017.
\newblock Accurate, large minibatch sgd: Training imagenet in 1 hour.
\newblock {\em arXiv preprint arXiv:1706.02677}.

\bibitem[\protect\citeauthoryear{He \bgroup et al\mbox.\egroup
  }{2016}]{he2016deep}
He, K.; Zhang, X.; Ren, S.; and Sun, J.
\newblock 2016.
\newblock Deep residual learning for image recognition.
\newblock In {\em CVPR}.

\bibitem[\protect\citeauthoryear{Howard \bgroup et al\mbox.\egroup
  }{2017}]{howard2017mobilenets}
Howard, A.~G.; Zhu, M.; Chen, B.; Kalenichenko, D.; Wang, W.; Weyand, T.;
  Andreetto, M.; and Adam, H.
\newblock 2017.
\newblock Mobilenets: Efficient convolutional neural networks for mobile vision
  applications.
\newblock {\em arXiv preprint arXiv:1704.04861}.

\bibitem[\protect\citeauthoryear{Hu, Shen, and Sun}{2018}]{hu2018squeeze}
Hu, J.; Shen, L.; and Sun, G.
\newblock 2018.
\newblock Squeeze-and-excitation networks.
\newblock In {\em Proceedings of the IEEE conference on computer vision and
  pattern recognition},  7132--7141.

\bibitem[\protect\citeauthoryear{Huang \bgroup et al\mbox.\egroup
  }{2017}]{huang2017densely}
Huang, G.; Liu, Z.; Van Der~Maaten, L.; and Weinberger, K.~Q.
\newblock 2017.
\newblock Densely connected convolutional networks.
\newblock In {\em Proceedings of the IEEE conference on computer vision and
  pattern recognition},  4700--4708.

\bibitem[\protect\citeauthoryear{Kingma and Ba}{2015}]{kingma2014adam}
Kingma, D.~P., and Ba, J.
\newblock 2015.
\newblock Adam: {A} method for stochastic optimization.
\newblock In Bengio, Y., and LeCun, Y., eds., {\em 3rd International Conference
  on Learning Representations, {ICLR}}.

\bibitem[\protect\citeauthoryear{Krizhevsky, Nair, and
  Hinton}{2009}]{krizhevsky2009cifar}
Krizhevsky, A.; Nair, V.; and Hinton, G.
\newblock 2009.
\newblock Cifar-10 and cifar-100 datasets.
\newblock {\em URl: https://www. cs. toronto. edu/kriz/cifar. html} 6.

\bibitem[\protect\citeauthoryear{Lan, Zhu, and Gong}{2018}]{lan2018self}
Lan, X.; Zhu, X.; and Gong, S.
\newblock 2018.
\newblock Self-referenced deep learning.
\newblock In {\em Asian Conference on Computer Vision},  284--300.
\newblock Springer.

\bibitem[\protect\citeauthoryear{Li \bgroup et al\mbox.\egroup
  }{2019}]{li2019stacnas}
Li, G.; Zhang, X.; Wang, Z.; Li, Z.; and Zhang, T.
\newblock 2019.
\newblock Stacnas: Towards stable and consistent optimization for
  differentiable neural architecture search.
\newblock {\em arXiv preprint arXiv:1909.11926}.

\bibitem[\protect\citeauthoryear{Liu \bgroup et al\mbox.\egroup
  }{2018}]{liu2018progressive}
Liu, C.; Zoph, B.; Neumann, M.; Shlens, J.; Hua, W.; Li, L.-J.; Fei-Fei, L.;
  Yuille, A.; Huang, J.; and Murphy, K.
\newblock 2018.
\newblock Progressive neural architecture search.
\newblock In {\em ECCV}.

\bibitem[\protect\citeauthoryear{Liu \bgroup et al\mbox.\egroup
  }{2019}]{liu2019auto}
Liu, C.; Chen, L.-C.; Schroff, F.; Adam, H.; Hua, W.; Yuille, A.~L.; and
  Fei-Fei, L.
\newblock 2019.
\newblock Auto-deeplab: Hierarchical neural architecture search for semantic
  image segmentation.
\newblock In {\em CVPR}.

\bibitem[\protect\citeauthoryear{Liu \bgroup et al\mbox.\egroup
  }{2020}]{liu2020autofis}
Liu, B.; Zhu, C.; Li, G.; Zhang, W.; Lai, J.; Tang, R.; He, X.; Li, Z.; and Yu,
  Y.
\newblock 2020.
\newblock Autofis: Automatic feature interaction selection in factorization
  models for click-through rate prediction.
\newblock {\em KDD}.

\bibitem[\protect\citeauthoryear{Liu, Simonyan, and Yang}{2019}]{liu2018darts}
Liu, H.; Simonyan, K.; and Yang, Y.
\newblock 2019.
\newblock {DARTS}: Differentiable architecture search.
\newblock In {\em ICLR}.

\bibitem[\protect\citeauthoryear{Luo \bgroup et al\mbox.\egroup
  }{2018}]{luo2018neural}
Luo, R.; Tian, F.; Qin, T.; Chen, E.; and Liu, T.-Y.
\newblock 2018.
\newblock Neural architecture optimization.
\newblock In {\em NeurIPS}.

\bibitem[\protect\citeauthoryear{Ma \bgroup et al\mbox.\egroup
  }{2018}]{ma2018shufflenet}
Ma, N.; Zhang, X.; Zheng, H.-T.; and Sun, J.
\newblock 2018.
\newblock Shufflenet v2: Practical guidelines for efficient cnn architecture
  design.
\newblock In {\em Proceedings of the European Conference on Computer Vision
  (ECCV)},  116--131.

\bibitem[\protect\citeauthoryear{Nayman \bgroup et al\mbox.\egroup
  }{2019}]{nayman2019xnas}
Nayman, N.; Noy, A.; Ridnik, T.; Friedman, I.; Jin, R.; and Zelnik{-}Manor, L.
\newblock 2019.
\newblock {XNAS:} neural architecture search with expert advice.
\newblock In {\em Advances in Neural Information Processing System, NeurIPS},
  1975--1985.

\bibitem[\protect\citeauthoryear{Noy \bgroup et al\mbox.\egroup
  }{2020}]{noy2019asap}
Noy, A.; Nayman, N.; Ridnik, T.; Zamir, N.; Doveh, S.; Friedman, I.; Giryes,
  R.; and Zelnik, L.
\newblock 2020.
\newblock Asap: Architecture search, anneal and prune.
\newblock In {\em International Conference on Artificial Intelligence and
  Statistics},  493--503.
\newblock PMLR.

\bibitem[\protect\citeauthoryear{Pham \bgroup et al\mbox.\egroup
  }{2018}]{pham2018efficient}
Pham, H.; Guan, M.~Y.; Zoph, B.; Le, Q.~V.; and Dean, J.
\newblock 2018.
\newblock Efficient neural architecture search via parameter sharing.
\newblock In {\em Proceedings of the 35th International Conference on Machine
  Learning, {ICML}}, volume~80 of {\em Proceedings of Machine Learning
  Research},  4092--4101.
\newblock {PMLR}.

\bibitem[\protect\citeauthoryear{Real \bgroup et al\mbox.\egroup
  }{2019}]{real2019regularized}
Real, E.; Aggarwal, A.; Huang, Y.; and Le, Q.~V.
\newblock 2019.
\newblock Regularized evolution for image classifier architecture search.
\newblock In {\em AAAI}.

\bibitem[\protect\citeauthoryear{Russakovsky \bgroup et al\mbox.\egroup
  }{2015}]{russakovsky2015ImageNet}
Russakovsky, O.; Deng, J.; Su, H.; Krause, J.; Satheesh, S.; Ma, S.; Huang, Z.;
  Karpathy, A.; Khosla, A.; Bernstein, M.; et~al.
\newblock 2015.
\newblock Imagenet large scale visual recognition challenge.
\newblock {\em International journal of computer vision} 115(3).

\bibitem[\protect\citeauthoryear{Sandler \bgroup et al\mbox.\egroup
  }{2018}]{sandler2018mobilenetv2}
Sandler, M.; Howard, A.; Zhu, M.; Zhmoginov, A.; and Chen, L.-C.
\newblock 2018.
\newblock Mobilenetv2: Inverted residuals and linear bottlenecks.
\newblock In {\em Proceedings of the IEEE Conference on Computer Vision and
  Pattern Recognition},  4510--4520.

\bibitem[\protect\citeauthoryear{Stamoulis \bgroup et al\mbox.\egroup
  }{2019}]{stamoulis2019single}
Stamoulis, D.; Ding, R.; Wang, D.; Lymberopoulos, D.; Priyantha, B.; Liu, J.;
  and Marculescu, D.
\newblock 2019.
\newblock Single-path nas: Designing hardware-efficient convnets in less than 4
  hours.
\newblock In {\em Joint European Conference on Machine Learning and Knowledge
  Discovery in Databases},  481--497.
\newblock Springer.

\bibitem[\protect\citeauthoryear{Szegedy \bgroup et al\mbox.\egroup
  }{2016}]{szegedy2016rethinking}
Szegedy, C.; Vanhoucke, V.; Ioffe, S.; Shlens, J.; and Wojna, Z.
\newblock 2016.
\newblock Rethinking the inception architecture for computer vision.
\newblock In {\em CVPR}.

\bibitem[\protect\citeauthoryear{Tan and Le}{2019}]{tan2019efficientnet}
Tan, M., and Le, Q.~V.
\newblock 2019.
\newblock Efficientnet: Rethinking model scaling for convolutional neural
  networks.
\newblock In Chaudhuri, K., and Salakhutdinov, R., eds., {\em Proceedings of
  the 36th International Conference on Machine Learning, {ICML}},  6105--6114.

\bibitem[\protect\citeauthoryear{Tan \bgroup et al\mbox.\egroup
  }{2019}]{tan2019mnasnet}
Tan, M.; Chen, B.; Pang, R.; Vasudevan, V.; Sandler, M.; Howard, A.; and Le,
  Q.~V.
\newblock 2019.
\newblock Mnasnet: Platform-aware neural architecture search for mobile.
\newblock In {\em CVPR}.

\bibitem[\protect\citeauthoryear{Xie \bgroup et al\mbox.\egroup
  }{2019}]{xie2018snas}
Xie, S.; Zheng, H.; Liu, C.; and Lin, L.
\newblock 2019.
\newblock {SNAS:} stochastic neural architecture search.
\newblock In {\em 7th International Conference on Learning Representations,
  {ICLR}}.

\bibitem[\protect\citeauthoryear{Xu \bgroup et al\mbox.\egroup
  }{2019a}]{hang2019auto-fpn}
Xu, H.; Yao, L.; Zhang, W.; Liang, X.; and Li, Z.
\newblock 2019a.
\newblock Auto-fpn: Automatic network architecture adaptation for object
  detection beyond classification.
\newblock In {\em ICCV}.

\bibitem[\protect\citeauthoryear{Xu \bgroup et al\mbox.\egroup
  }{2019b}]{xu2019pc}
Xu, Y.; Xie, L.; Zhang, X.; Chen, X.; Qi, G.-J.; Tian, Q.; and Xiong, H.
\newblock 2019b.
\newblock Pc-darts: Partial channel connections for memory-efficient
  differentiable architecture search.
\newblock {\em arXiv preprint arXiv:1907.05737}.

\bibitem[\protect\citeauthoryear{Yang, Esperan{\c{c}}a, and
  Carlucci}{2020}]{yang2019evaluation}
Yang, A.; Esperan{\c{c}}a, P.~M.; and Carlucci, F.~M.
\newblock 2020.
\newblock Nas evaluation is frustratingly hard.
\newblock {\em ICLR 2020}.

\bibitem[\protect\citeauthoryear{Yao \bgroup et al\mbox.\egroup
  }{2019}]{yao2019differentiable}
Yao, Q.; Xu, J.; Tu, W.-W.; and Zhu, Z.
\newblock 2019.
\newblock Differentiable neural architecture search via proximal iterations.
\newblock {\em arXiv preprint arXiv:1905.13577}.

\bibitem[\protect\citeauthoryear{Zela \bgroup et al\mbox.\egroup
  }{2020}]{zela2019understanding}
Zela, A.; Elsken, T.; Saikia, T.; Marrakchi, Y.; Brox, T.; and Hutter, F.
\newblock 2020.
\newblock Understanding and robustifying differentiable architecture search.
\newblock {\em ICLR}.

\bibitem[\protect\citeauthoryear{Zhang \bgroup et al\mbox.\egroup
  }{2018}]{zhang2017mixup}
Zhang, H.; Ciss{\'{e}}, M.; Dauphin, Y.~N.; and Lopez{-}Paz, D.
\newblock 2018.
\newblock mixup: Beyond empirical risk minimization.
\newblock In {\em 6th International Conference on Learning Representations,
  {ICLR}}.

\bibitem[\protect\citeauthoryear{Zhou \bgroup et al\mbox.\egroup
  }{2020}]{zhou2020theory}
Zhou, P.; Xiong, C.; Socher, R.; and Hoi, S.~C.
\newblock 2020.
\newblock Theory-inspired path-regularized differential network architecture
  search.
\newblock {\em Advances in Neural Information Processing System, NeurIPS}.

\bibitem[\protect\citeauthoryear{Zoph and Le}{2017}]{regular2016neural}
Zoph, B., and Le, Q.~V.
\newblock 2017.
\newblock Neural architecture search with reinforcement learning.
\newblock In {\em 5th International Conference on Learning Representations,
  {ICLR}}.

\bibitem[\protect\citeauthoryear{Zoph \bgroup et al\mbox.\egroup
  }{2018}]{zoph2018learning}
Zoph, B.; Vasudevan, V.; Shlens, J.; and Le, Q.~V.
\newblock 2018.
\newblock Learning transferable architectures for scalable image recognition.
\newblock In {\em CVPR}.

\end{thebibliography}
\clearpage

\appendix

\section{Proof of Lemma 1}\label{proof-lemma}

{\em Proof.} \textbf{P1}: denote $\alpha_1 = 1 - \alpha_0$ and $\mathbf{W}_\alpha = \alpha_0 \mathbf{I} + \alpha_1 \mathbf{W}$, the gradients of $\mathbf{W}, \mathbf{w}_r$ regarding to the training loss $\mathcal{L}_{train}$ is 
\begin{equation*}
\begin{split}
\nabla_{\mathbf{w}_r} \mathcal{L}_{train} &= \sum_{(\mathbf{x}_i, y_i) \in \mathcal{T}} [\sigma(y_i \mathbf{w}_r^\top \mathbf{W}_\alpha \mathbf{x}_i) -1] y_i \mathbf{W}_\alpha \mathbf{x}_i, \\
\nabla_\mathbf{W} \mathcal{L}_{train} &= \sum_{(\mathbf{x}_i, y_i) \in \mathcal{T}} [\sigma(y_i \mathbf{w}_r^\top \mathbf{W}_\alpha \mathbf{x}_i) -1] y_i \alpha_1 \mathbf{w}_r \mathbf{x}_i^\top,
\end{split}
\end{equation*}
where $\sigma(x) = \frac{1}{1 + \exp(-x)}$ is the sigmoid function. 

Suppose that $y \mathbf{x} = \mu_t \mathbf{e} + \sigma_t \mathbf{\epsilon}, (\mathbf{x}, y) \in \mathcal{T}, \mathbf{\epsilon} \sim N(0, \mathbf{I})$. Denote $\mathbf{v} = \mathbf{W}_\alpha^\top \mathbf{w}_r$ and $\mathbf{v}_\perp$ holds such that $\| \mathbf{v}_\perp \| = \|\mathbf{v}\|$ and $ \mathbf{v}_\perp^\top \mathbf{v} = 0$. As $\mathbf{\epsilon}$ is isotropic, $\mathbf{\epsilon}$ can be written as $\sigma_t \mathbf{\epsilon} = \sigma_t' (\mathbf{v} \epsilon_0 + \mathbf{v}_\perp \epsilon_1), \epsilon_0, \epsilon_1 \sim N(0,1) $ where $\sigma_t' = \sigma_t / \| \mathbf{v} \|$. Note that the training error is expected to be small, which corresponds to smaller $\sigma_t$. Then the gradients of $\mathbf{w}_r$ can be rewritten as 
\begin{equation*}
\begin{split}
\nabla_{\mathbf{w}_r} \mathcal{L}_{train} =& \mathbb{E}_{\epsilon_0, \epsilon_1} \big\{ [ \sigma(\mu_t \mathbf{v}^\top \mathbf{e} + \sigma_t' \mathbf{v}^\top (\mathbf{v} \epsilon_0 + \mathbf{v}_\perp \epsilon_1))-1] \\
&[\mu_t \mathbf{W}_\alpha \mathbf{e} + \sigma_t' \mathbf{W}_\alpha (\mathbf{v} \epsilon_0 + \mathbf{v}_\perp \epsilon_1) ] \big\} \\
=& \mu_t \mathbb{E}_{\epsilon_0} \big[ \sigma(\mu_t \mathbf{v}^\top \mathbf{e} + \sigma_t' \|\mathbf{v}\|^2 \epsilon_0)-1 \big] (\mathbf{W}_\alpha \mathbf{e}) \\
&+ \sigma_t' \mathbb{E}_{\epsilon_0, \epsilon_1} \big\{ [ \sigma(\mu_t \mathbf{v}^\top \mathbf{e} + \sigma_t \|\mathbf{v}\| \epsilon_0)-1 ] \epsilon_0 \\
&\qquad \qquad \qquad \cdot(\mathbf{W}_\alpha \mathbf{v} + \mathbf{W}_\alpha \mathbf{v}_\perp \epsilon_1) \big\} \\
=& \lambda_1 \mathbf{W}_\alpha \mathbf{e} + \lambda_2 \mathbf{W}_\alpha \mathbf{v}.
\end{split}
\end{equation*}
The last equality holds such that $\lambda_1 = \mu_t \mathbb{E}_{\epsilon_0} \big[ \sigma(\mu_t \mathbf{v}^\top \mathbf{e} + \sigma_t \|\mathbf{v} \| \epsilon_0)-1 \big], \lambda_2 = \sigma_t' \mathbb{E}_{\epsilon_0} \big\{ [ \sigma(\mu_t \mathbf{v}^\top \mathbf{e} + \sigma_t \|\mathbf{v}\| \epsilon_0)-1 ] \epsilon_0 \}$. Consider $\sigma_t$ is small underlying the training data and $\| \mathbf{v} \|$ is limited as $\{ \mathbf{W} \mathbf{x}, \mathbf{x} \in \mathcal{T} \}$ is normalized, it is expected that $|\lambda_1| \gg |\lambda_2|$. Thus we have $\nabla_{\mathbf{w}_r} \mathcal{L}_{train} \approx \lambda_1 \mathbf{W}_\alpha \mathbf{e}$. If $\mathbf{w}_r$ is optimized with $\mathbf{W}$ fixed, we have $\mathbf{w}_r \propto \mathbf{W}_\alpha \mathbf{e}$ with gradient descent. $\propto$ denotes one vector/matrix is parallel to the other, where $\mathbf{a} \propto \mathbf{b} $ implies there exists $\psi \neq 0$ such that $\mathbf{a} = \psi \mathbf{b}$.

The gradients of $\mathbf{W}$ is 
\begin{equation*}
\begin{split}
\nabla_{\mathbf{W}} \mathcal{L}_{train} =& \mathbb{E}_{\epsilon_0, \epsilon_1} \big\{ [ \sigma(\mu_t \mathbf{v}^\top \mathbf{e} + \sigma_t' \mathbf{v}^\top (\mathbf{v} \epsilon_0 + \mathbf{v}_\perp \epsilon_1))-1] \\
&\alpha_1 [\mu_t \mathbf{w}_r \mathbf{e}^\top + \sigma_t' \mathbf{w}_r (\mathbf{v} \epsilon_0 + \mathbf{v}_\perp \epsilon_1) ]^\top \big\} \\
=& \alpha_1 [ \lambda_1 \mathbf{w}_r \mathbf{e}^\top + \lambda_2 \mathbf{w}_r \mathbf{v}^\top].
\end{split}
\end{equation*}
Similar as above, it is expected that $|\lambda_1| \gg |\lambda_2|$ and therefore $\nabla_{\mathbf{W}} \mathcal{L}_{train} \approx \lambda_1 \mathbf{w}_r \mathbf{e}^\top$, thus $\mathbf{W} \propto \alpha_1 \mathbf{w}_r \mathbf{e}^\top$ with gradient descent. 

Then we have $\mathbf{w}_r \propto \mathbf{W}_\alpha \mathbf{e} = \alpha_0 \mathbf{e} + \alpha_1 \mathbf{W} \mathbf{e} = \alpha_0 \mathbf{e} + \gamma_1 \mathbf{w}_r \mathbf{e}^\top \mathbf{e} = \gamma_0 \mathbf{e} + \gamma_1 \mathbf{w}_r$ where $\gamma_0, \gamma_1$ are certain constants. Thus during the iteration, it is expected that $\mathbf{w}_r \propto \mathbf{e}$ and therefore $\mathbf{W} \to \mathbf{W}^* \propto \mathbf{e} \mathbf{e}^\top$.

As $\| \mathbf{w}_r \| = r$, we have $\mathbf{w}_r = r \mathbf{e}$.

Denote $\mathbf{W}^* = \eta \mathbf{e} \mathbf{e}^\top = \frac{\eta}{2} \mathbf{1}_{2 \times 2}$. As $\{\mathbf{W} \mathbf{x}, \mathbf{x} \in \mathcal{T}\}$ is normalized, the variance of first dimension is $\mathbb{D}_{\mathcal{T}} [(\mathbf{W} \mathbf{x}) [0]] = \mathbb{D}_{\mathcal{T}} [\frac{\eta}{2}(\mathbf{x}[0] + \mathbf{x}[1])] = \frac{\eta^2}{4} \mathbb{E}_{(\mathbf{x}, y) \sim \mathcal{T}} [\mathbf{x}[0]^2 + \mathbf{x}[1]^2 + 2(y \mathbf{x}[0]) (y \mathbf{x}[1])] = \frac{\eta^2}{4} (1 + 1 + 2\cdot \frac{\sqrt{2}}{2} \mu_t \cdot \frac{\sqrt{2}}{2} \mu_t) = \frac{\eta^2 (2 + \mu_t^2)}{4}$ and so as $\mathbb{D}_{\mathcal{T}} [(\mathbf{W} \mathbf{x}) [1]] = \frac{\eta^2 (2 + \mu_t^2)}{4}$. As the variance of each dimension should be 1, we have $\eta = \frac{2}{\sqrt{2 + \mu_t^2}}$. Then $\mathbf{W}^* \mathbf{x} = (\eta \mathbf{e}^\top \mathbf{x}) \mathbf{e}$, which completes the proof of \textbf{P1}.

\textbf{P2}. The gradient of $\alpha_0$ regarding to $\mathcal{L}_{val}$ is
\begin{equation*}
\begin{split}
\frac{\mathrm{d} \mathcal{L}_{val}}{\mathrm{d} \alpha_0} = \sum_{(\mathbf{x}_i, y_i) \in \mathcal{V}} [\sigma(y_i \mathbf{w}_r^\top \mathbf{W}_\alpha \mathbf{x}_i) -1] y_i (\mathbf{w}_r^\top \mathbf{x}_i - \mathbf{w}_r^\top \mathbf{W} \mathbf{x}_i).
\end{split}
\end{equation*}
When \textbf{P1} holds, $\mathbf{w}_r = r \mathbf{e}, \mathbf{W} = \frac{2}{\sqrt{2 + \mu_t^2}} \mathbf{e} \mathbf{e}^\top$, $\mathbf{W}_\alpha^\top \mathbf{w}_r = r (\alpha_0 + \frac{2\alpha_1}{\sqrt{2 + \mu_t^2}}) \mathbf{e}$. Denote $\lambda = \alpha_0 + \frac{2\alpha_1}{\sqrt{2 + \mu_t^2}}$ and $\mathbf{x} = \mu_v \mathbf{e} + \sigma_v \epsilon, \epsilon \sim N(0, \mathbf{I})$, we have 
\begin{equation*}
\begin{split}
\frac{\mathrm{d} \mathcal{L}_{val}}{\mathrm{d} \alpha_0} =& \sum_{(\mathbf{x}_i, y_i) \in \mathcal{V}} [\sigma(r \lambda y_i \mathbf{e}^\top \mathbf{x}_i) -1] r y_i (1-\lambda) \mathbf{e}^\top \mathbf{x}_i \\
=& \mathbb{E}_{\epsilon} \big\{ [\sigma(r \lambda (\mu_v + \sigma_v \mathbf{e}^\top \epsilon)) - 1] r (1-\lambda) (\mu_v + \sigma_v \mathbf{e}^\top \epsilon) \big\} \\
=& r (1-\lambda) \mathbb{E}_{\epsilon_1} \big\{ [\sigma(r \lambda (\mu_v + \sigma_v \epsilon_1)) - 1] (\mu_v + \sigma_v \epsilon_1) \big\}.
\end{split}
\end{equation*}
The last equality holds such that $\epsilon = \epsilon_1 \mathbf{e} + \epsilon_2 \mathbf{e}_\perp, \epsilon_1, \epsilon_2 \sim N(0,1)$ where $\mathbf{e}_\perp = \frac{1}{\sqrt{2}}[-1,1]^\top$, in this case $\mathbf{e}^\top \epsilon = \epsilon_1$. Moreover, as $r > 0, \mu_t \in [0, \sqrt{2}]$, $1 - \lambda \le 1 - (\alpha_0 + \alpha_1) = 0$ and the equality only holds when $\mu_t = \sqrt{2}$ or $\sigma_t = 0$.

Consider that $\frac{1}{2} \mu_v^2 + \sigma_v^2 = 1$, we have $\mu_v \in [0, \sqrt{2}]$ and $\sigma_v \in [0, 1]$. Denote 
\begin{equation*}
g(r, \sigma_v) = - \mathbb{E}_{\epsilon_1} \big\{ [\sigma(r \lambda (\mu_v + \sigma_v \epsilon_1)) - 1] (\mu_v + \sigma_v \epsilon_1) \big\}.
\end{equation*}
We will show there exists $\sigma_0 (r) \in (0,1)$ such that $g(r, \sigma_v ) < 0, \sigma_v > \sigma_0 (r) $. It is clear that $g(\sigma_v)$ is continuous at $[0,1]$. $g(r, 0) = - \sqrt{2} [\sigma(\sqrt{2} r \lambda) - 1] > 0$. While $g(r, 1) = - \mathbb{E}_{\epsilon_1} \big\{ [\sigma(r \lambda \epsilon_1) - 1]  \epsilon_1 \big\} = -\frac{1}{2} \mathbb{E}_{\epsilon_1} \big\{ [\sigma(r \lambda \epsilon_1) - 1]  \epsilon_1 \big\} - \frac{1}{2} \mathbb{E}_{\epsilon_1} \big\{ [\sigma(-r \lambda \epsilon_1) - 1]  (-\epsilon_1) \big\} = \frac{1}{2} \mathbb{E}_{\epsilon_1} \big\{ [\sigma(-r \lambda \epsilon_1) - \sigma(r \lambda \epsilon_1)]  \epsilon_1 \big\} < 0$ as $[\sigma(-r \lambda \epsilon_1) - \sigma(r \lambda \epsilon_1)]  \epsilon_1 \le 0$ for any $r, \epsilon_1$. As $g(r, 1) < 0 < g(r, 0)$ and $g(r, \sigma_v)$ is continuous for any $r > 0$, by the Intermediate Value Theorem for continuous function, there exists $\sigma_0 (r) \in (0,1)$ such that $g(r, \sigma_0 (r)) = 0$ and $g(r, \sigma_v) < 0$ if $\sigma_v > \sigma_0 (r) $. Thus $\frac{\mathrm{d} \mathcal{L}_{val}}{\mathrm{d} \alpha_0} < 0$ if $\sigma_v > \sigma_0 (r) $ as $1-\lambda < 0$, which means that $\alpha_0$ will get larger with gradient descent. 

Moreover, The partial derivative of $g(r, \sigma_v)$ regarding to $r$ is
\begin{equation*}
\begin{split}
\frac{\partial g(r, \sigma_v)}{\partial r} &= - \mathbb{E}_{\epsilon_1} \big\{ \frac{\partial}{\partial r}[\sigma(r \lambda (\mu_v + \sigma_v \epsilon_1)) - 1] (\mu_v + \sigma_v \epsilon_1) \big\} \\
&= - \mathbb{E}_{\epsilon_1} \big\{ \sigma(\cdot) (1 - \sigma(\cdot) ) (\mu_v + \sigma_v \epsilon_1)^2 \big\} \\
&< 0.
\end{split}
\end{equation*}
where $\sigma(\cdot)$ denotes $\sigma(r \lambda (\mu_v + \sigma_v \epsilon_1))$. It is clear that $g(r, \sigma_v)$ is momentously decreasing for any $\sigma_v \in [0,1]$. Consider $r_1, r_2 > 0$, if there exists $\sigma_0(r_1)$ such that $g(r_1, \sigma_0 (r_1)) = 0$ and $g(r_1, \sigma_v) < 0$ if $\sigma_v > \sigma_0 (r_1) $, it is clear that $g(r_2, \sigma_0 (r_1)) < 0$. With the Intermediate Value Theorem, there exists $\sigma_0(r_2) \in (0, \sigma_0(r_1))$ such that $g(r_2, \sigma_0 (r_2)) = 0$ and $g(r_2, \sigma_v) < 0$ if $\sigma_v > \sigma_0 (r_2)$. Therefore, $\sigma_0 (r_2) < \sigma_0 (r_1)$ holds for any $r_2 > r_1 > 0$, thus $\sigma_0 (r)$ is a momentously decreasing function. Then the proof is completed.

\begin{figure}
    \centering
    \includegraphics[width=3in]{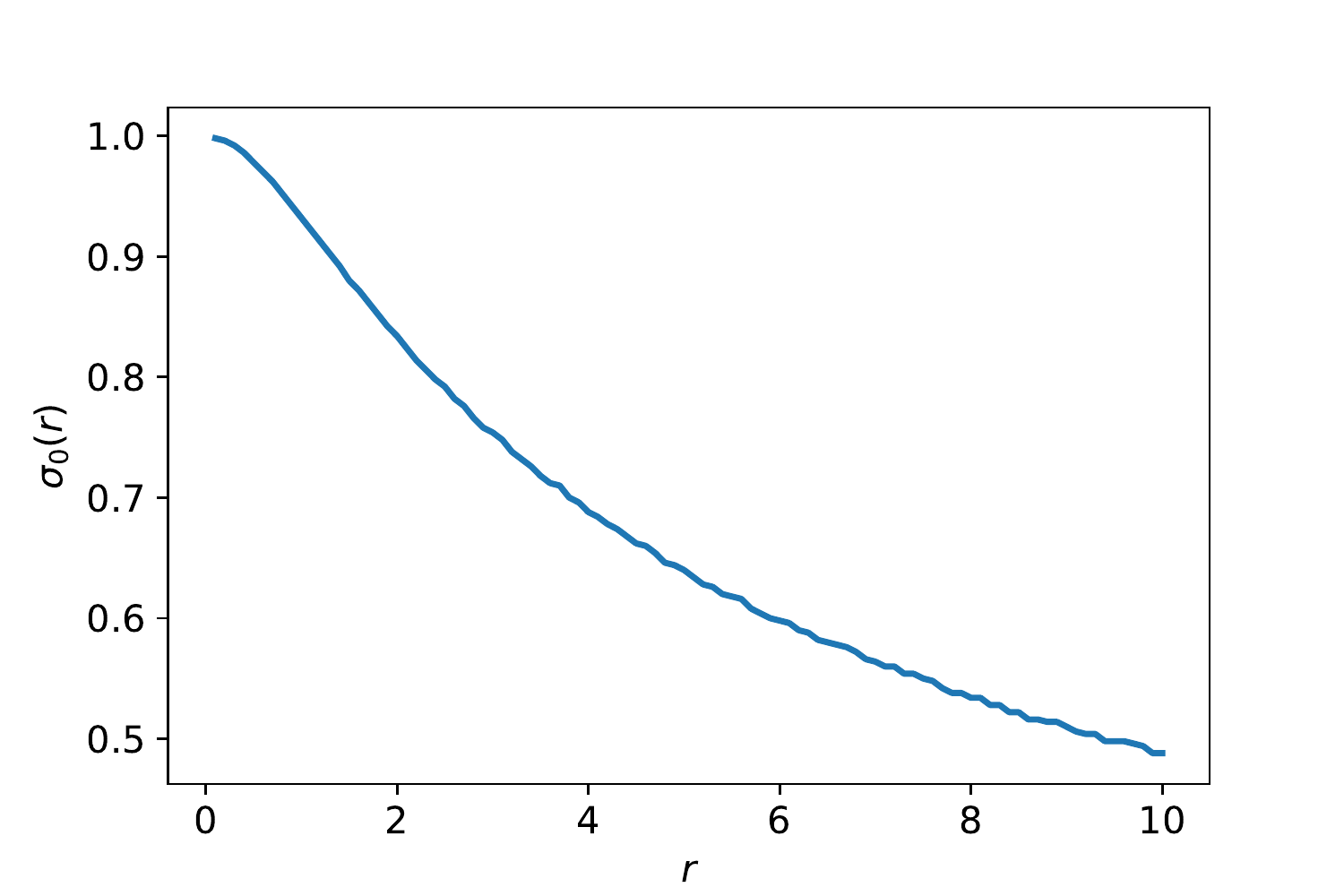}
    \caption{The numerical value of $\sigma_0 (r)$.}
    \label{fig:sigma_r}
\end{figure}

\textbf{Remark} Fig. \ref{fig:sigma_r} shows the numerical value of $\sigma_0 (r)$ when $\alpha_0 = 0.5$. In fact, $\sigma_0 (r)$ is not sensitive to $\alpha_0$. 

\begin{figure*}[t]
    \setlength{\abovecaptionskip}{0pt}
    \setlength{\belowcaptionskip}{0pt}
    	\begin{center}
    		\includegraphics[width=1.00\linewidth]{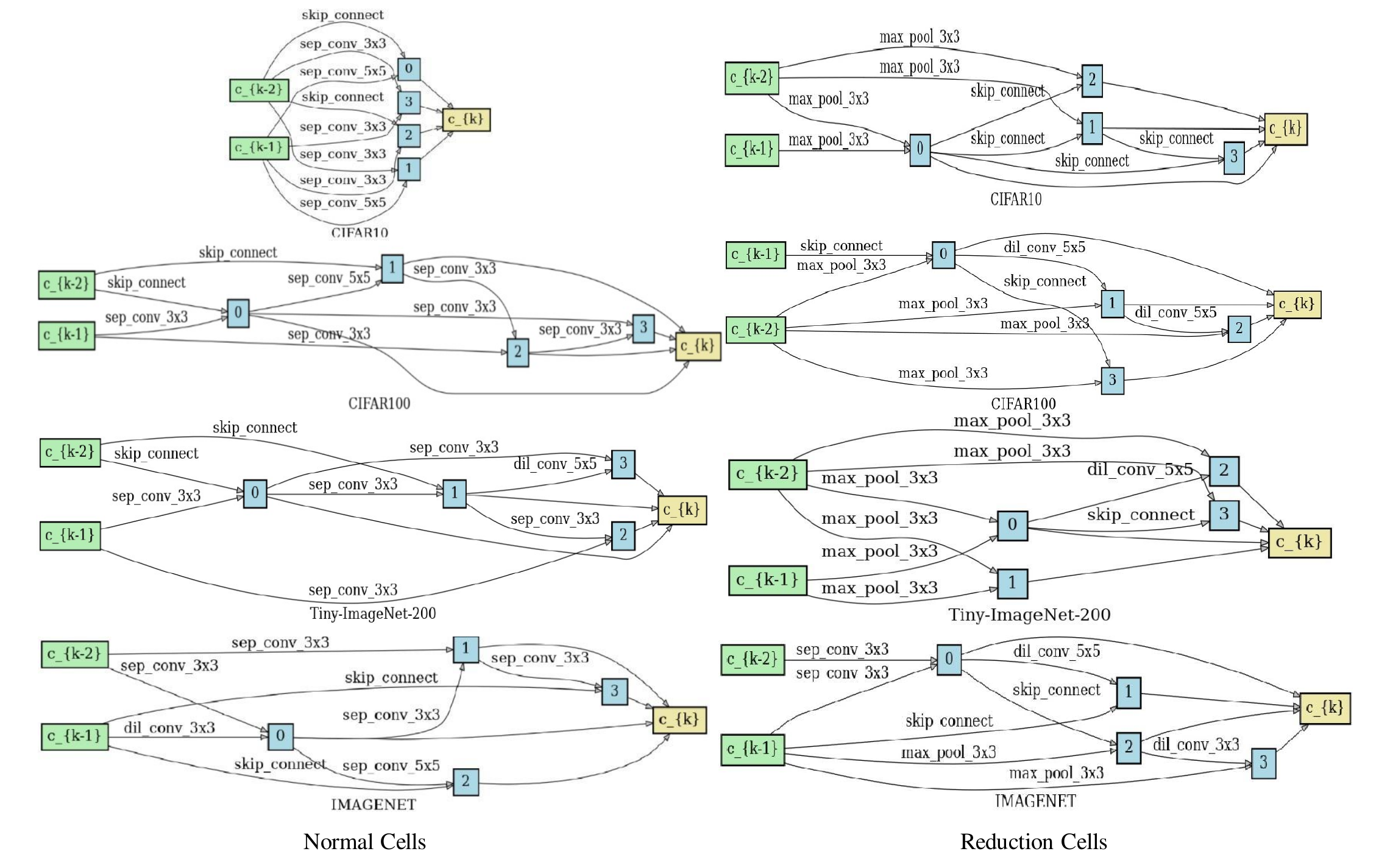}
    		\caption{The cell of best structures searched on different datasets.}
    		\label{best_struc}
    	\end{center}
    \end{figure*}

\section{Basic Experimental Settings}

\subsection{Architecture Search}

\subsubsection{DARTS Search Space.}
Our search space on DARTS is the same as original DARTS~\cite{liu2018darts} which has 8 candidate operations including {\em skip-connect, max-pool-3x3, avg-pool-3x3, sep-conv-3x3, sep-conv-5x5, dil-conv-3x3, dil-conv-5x5, zero}, and the structure of each operation is exactly the same as DARTS.

For CIFAR10 and CIFAR100, we use the same one-shot model as the original DARTS in which 8 cells (i.e. 6 normal cells and 2 reduction cells) with 16 channels are trained for searching. We use half of the training data to train the model weights and the other half to update the architecture parameters. We search for a maximum of $60$ epochs with batch size $64$. SGD is used to optimize the model weights with initial learning rate $0.025$, momentum $0.9$, and weight decay $3\times10^{-4}$. Adam~\cite{kingma2014adam} is adopted to optimize architecture parameters with initial learning rate $3\times10^{-4}$, momentum $(0.5,0.999)$ and weight decay $10^{-3}$. Early stopping is applied at a certain epoch when Criterion 1 is met.

For Tiny-ImageNet-200, the one-shot model is almost the same as CIFAR10/100 except that a $3\times3$ convolution layer with stride 2 is added on the first layer to reduce the input resolution from $64 \times 64$ to $32 \times 32$. Other settings are the same as those used in CIFAR10/100, including the ``early stopping'' criterion.

For ImageNet, following~\cite{xu2019pc}, the one-shot model starts with three $3\times 3$ convolution layers with stride 2 to reduce the resolution from $224 \times 224$ to $28 \times 28$, and the rest of the network consists of $8$ cells. We select $10\%$ data from the training set for updating model weights and another $10\%$ for updating architecture parameters. We search with batch size $512$ for both training and validation sets. SGD is used for model weights training with initial learning rate $0.2$ (cosine learning rate decay), momentum $0.9$, and weight decay $3\times 10^{-4}$. The architecture parameters are trained with Adam with learning rate $3\times 10^{-3}$, momentum $(0.5, 0.999)$ and weight decay $10^{-3}$.

For all the datasets, the one-shot model weights and architecture parameters are optimized alternatively. The cell structure is determined by architecture parameters, following DARTS~\cite{liu2018darts}.

The selected architectures are shown in Fig. \ref{best_struc}. We observe that the cells searched by DARTS+ contain most {\em convolution}s and a few {\em skip-connect}s. 

\begin{table}[t]
    	\caption{The MobileNetV2 backbone model used for searching and evaluation. $c, n, s$ are the same as that in \cite{sandler2018mobilenetv2}.}
    	\centering
    	\small
    	\begin{tabular}{ccccc}
    		\hline
    		Input & Operator & $c$ & $n$ & $s$ \\
    		\hline
            $32^2 \times 3$ & conv2d & 32 & 1 & 1 \\
            $32^2 \times 32$ & bottleneck & 16 & 1 & 1 \\
            $32^2 \times 16$ & bottleneck & 24 & 4 & 2 \\
            $16^2 \times 24$ & bottleneck & 40 & 4 & 2 \\     $8^2 \times 40$ & bottleneck & 80 & 4 & 2 \\      $4^2 \times 80$ & bottleneck & 96 & 4 & 1 \\      $4^2 \times 96$ & bottleneck & 192 & 4 & 1 \\     $4^2 \times 192$ & bottleneck & 320 & 1 & 1 \\  
            $4^2 \times 320$ & avgpool 4x4 & - & 1 & - \\  
            $1^2 \times 320$ & conv2d 1x1 & k & 1 & - \\  
    	\hline
    	\end{tabular}
    	\label{tab:mobilenetv2_arch}
\end{table}

\begin{figure*}[h]
    \begin{center}
        \includegraphics[width=7in]{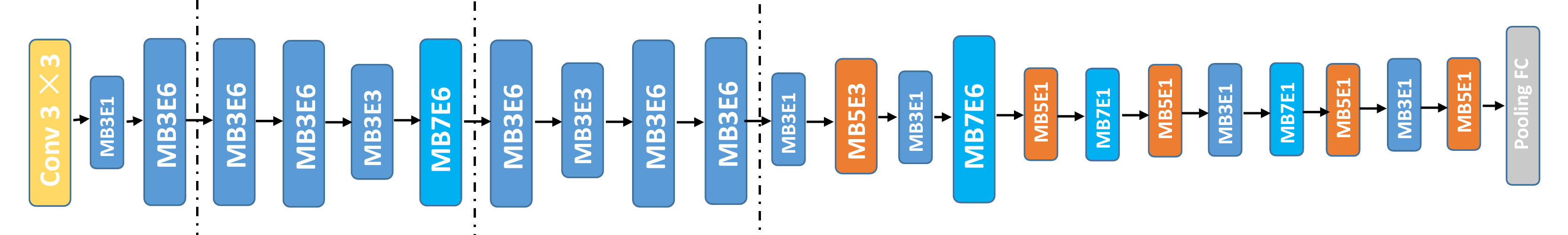}
        \caption{Architecture selected by DARTS+ in MobileNet-V2 search space on CIFAR100.}
        \label{mvb2_architect}
    \end{center}
\end{figure*}

\begin{figure*}[h]
    \begin{center}
        \includegraphics[width=7in]{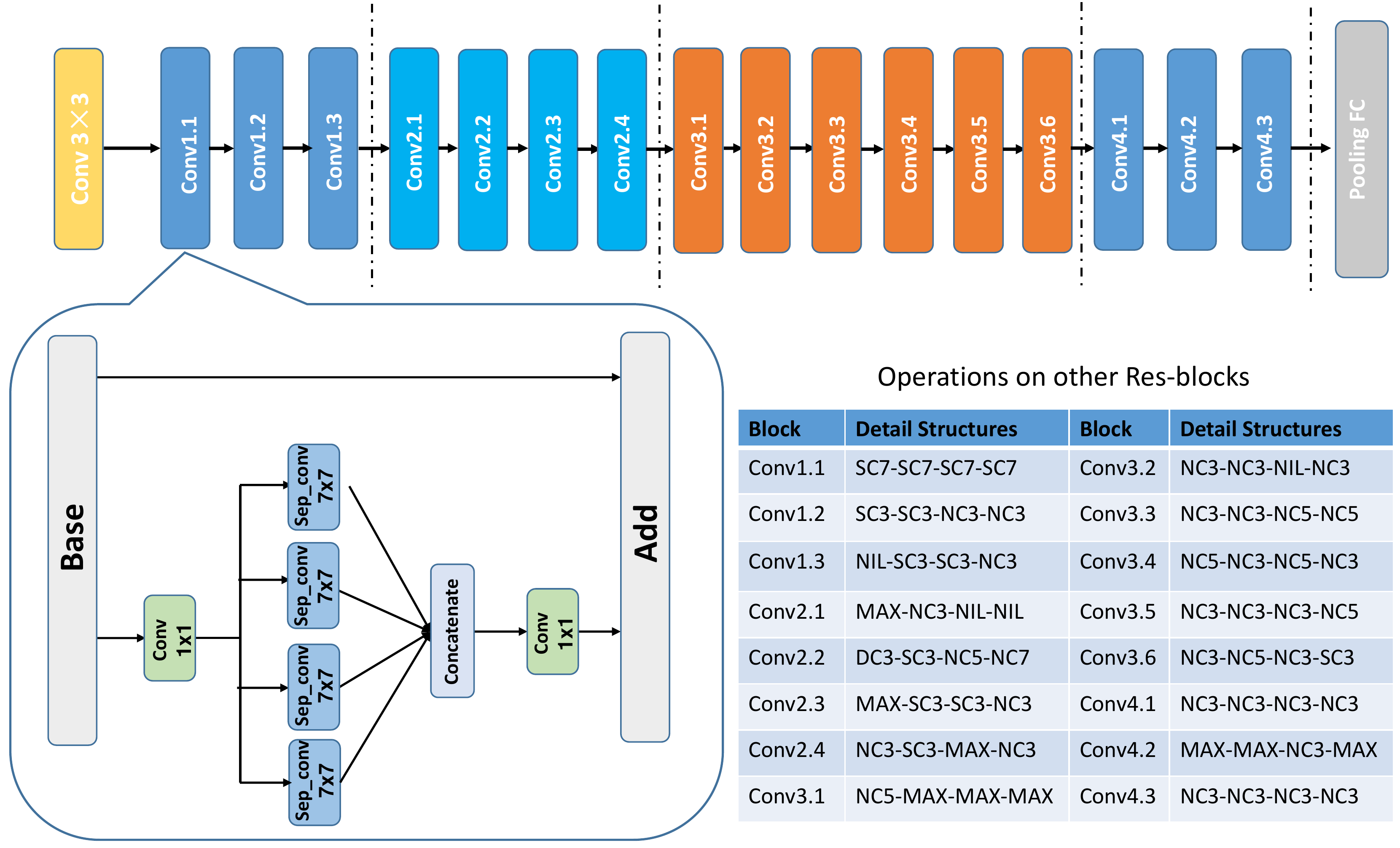}
        \caption{Architecture selected by DARTS+ in ResNet based search space on CIFAR100. ``SC'' denotes separable convolution, ``NC'' denotes normal convolution, ``MAX'' denotes max pooling layer, ``NIL'' denotes zero operation, ``DC'' denotes dilated convolution.}
        \label{res_architect}
    \end{center}
\end{figure*}

\subsubsection{MobileNetV2 Search Space.} 

In this search space, CIFAR100 is used for searching and evaluating. We follow~\cite{cai2018proxylessnas} and use MobileNetV2~\cite{sandler2018mobilenetv2} as the backbone to construct architecture space. We initiate each layer of the one-shot model by a set of mobile inverted bottleneck convolution(MBConv) layers with various kernel sizes \{3, 5, 7\}, and expansion rations \{3, 6\}. As the input image size is just $32\times 32$, The backbone is slightly different from the original MobielNetV2, shown in Fig. \ref{tab:mobilenetv2_arch}. Different from ~\cite{cai2018proxylessnas} which uses the binary gate to connect operations within each layer, we follow DARTS and apply softmax to architecture parameters to get weights of candidate operations. The architecture is determined by choosing operations with the largest architecture parameter $\alpha$ in each layer. We search the model on CIFAR100 for 50 epochs and perform early stopping with Criterion 2, which is based on the rank of last layer architecture parameters. 

The selected architecture with early stopping Criterion 2 is shown in Fig. \ref{mvb2_architect}. 

% \begin{table}[t]
%         \setlength{\abovecaptionskip}{0pt}
%         \setlength{\belowcaptionskip}{0pt}
%     	\caption{Results of different architectures searched from MobileNet-V2 and ResNet-50 backbone based model.}
%     	\centering
%     	\small
%     	\begin{tabular}{ccc}
%     		\hline
%     		Architecture backbone & MobileNet-V2 & ResNet-50 \\
%     		\hline
%     		Original & 70.08 & 80.26 \\
%     		Before EarlyStop & $70.87$ & $79.13$ \\
%     		\textbf{DARTS+} & $\mathbf{72.43}$ & $\mathbf{81.23}$ \\
%     		After EarlyStop & $71.56$ & $79.56$ \\
%     	\hline
%     	\end{tabular}
%     	\label{tab:supp_result}
% \end{table}

% \begin{figure*}[h]
%     \begin{center}
%         \includegraphics[width=7in]{imgs/mbv2_resnet_deepest.pdf}
%         \caption{Architecture parameters of the last searching layer of MobileNet-V2 and ResNet-50 based one-shot model. The early stopping points with Criterion 2 are marked as red circle.}
%         \label{mvb2_final}
%     \end{center}
% \end{figure*}

\begin{figure*}[htbp]
    \centering
    \includegraphics[width=5.5in]{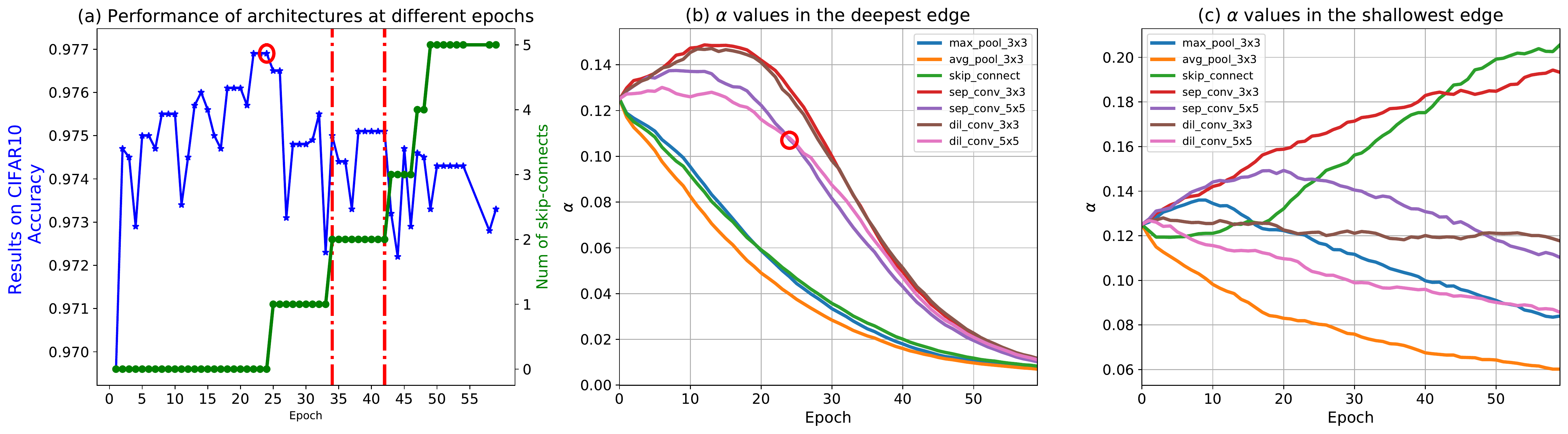} \par
    \includegraphics[width=5.5in]{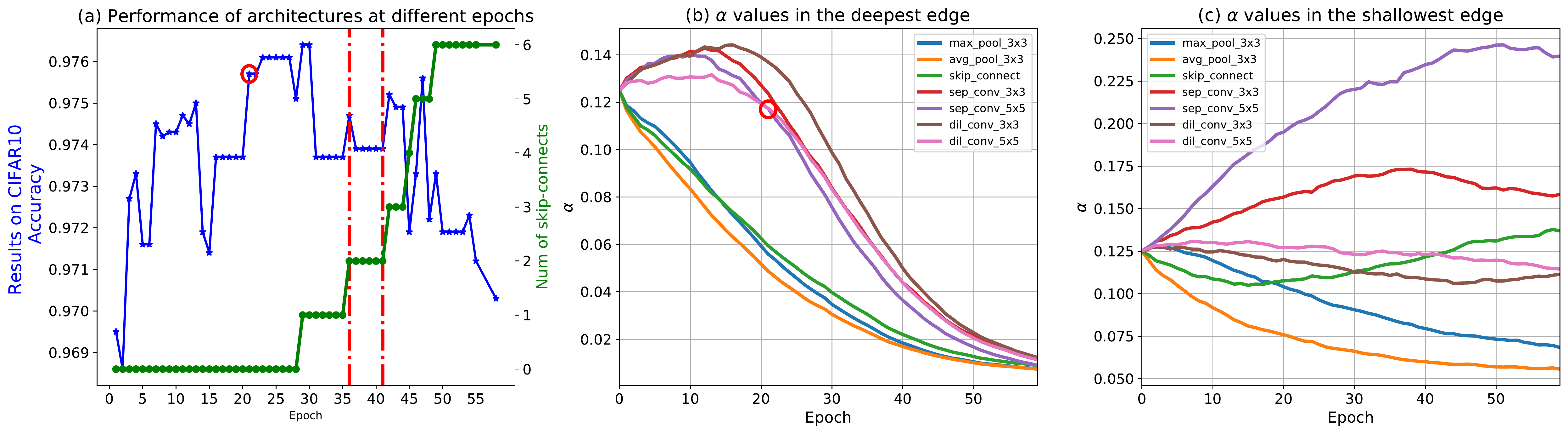} \par
    \caption{More results on CIFAR10 dataset. The meaning of each figure is the same as that in Fig. 2 in the main paper.}
    \label{all_in_three_cifar10}
\end{figure*}

\begin{figure*}[htbp]
    \centering
    \includegraphics[width=5.5in]{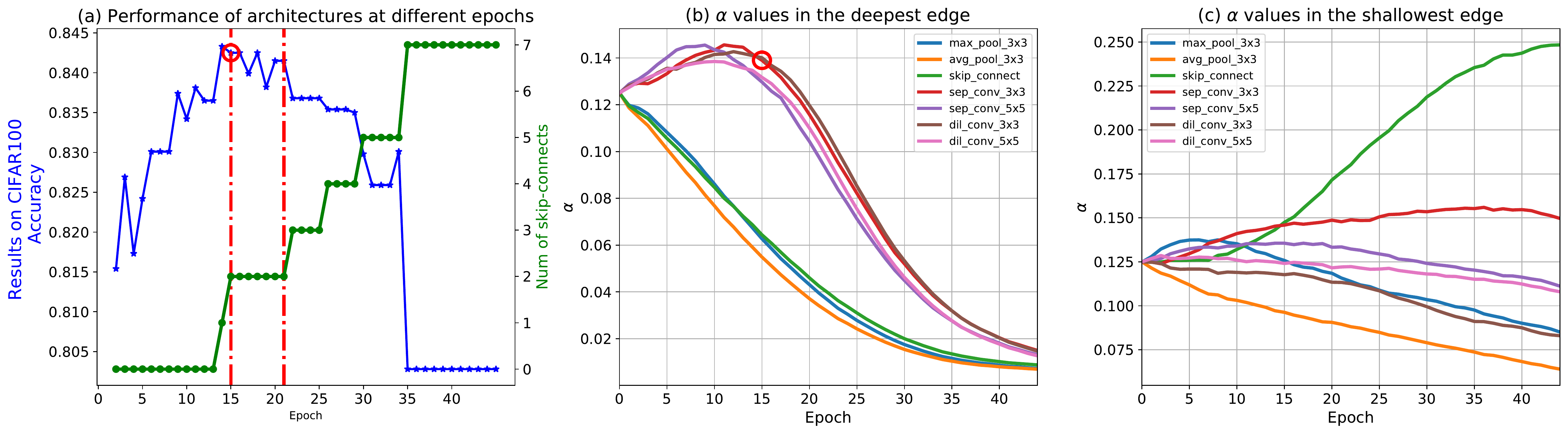} \par
    \includegraphics[width=5.5in]{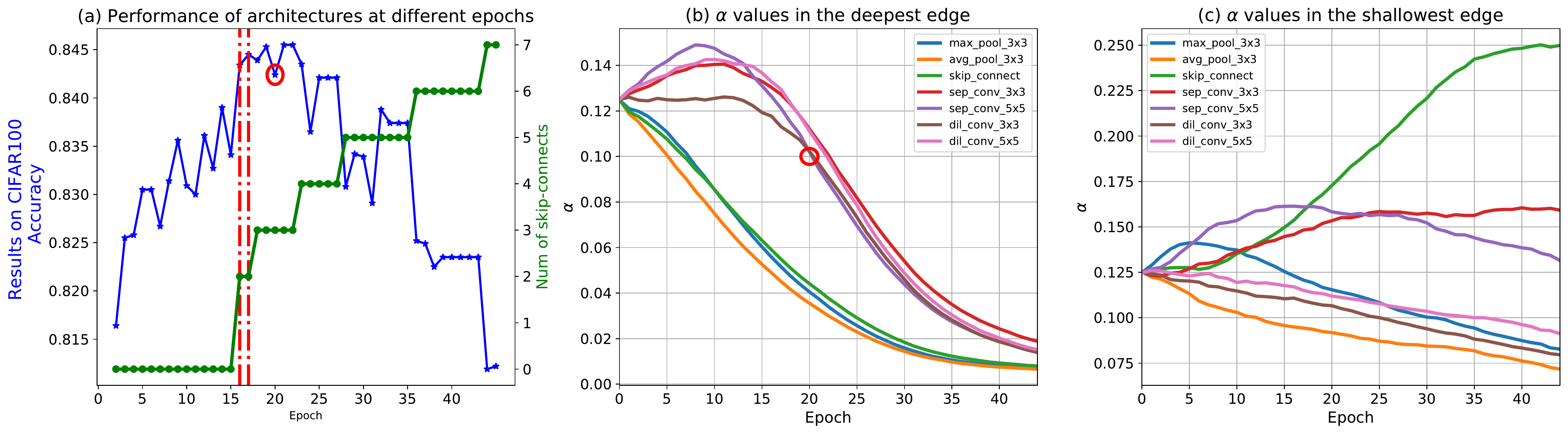} \par
    \caption{More results on CIFAR100 dataset. The meaning of each figure is the same as that in Fig. 2 in the main paper.}
    \label{all_in_three_cifar100}
\end{figure*}

\begin{figure*}[htbp]
    \begin{center}
        \includegraphics[width=5.5in]{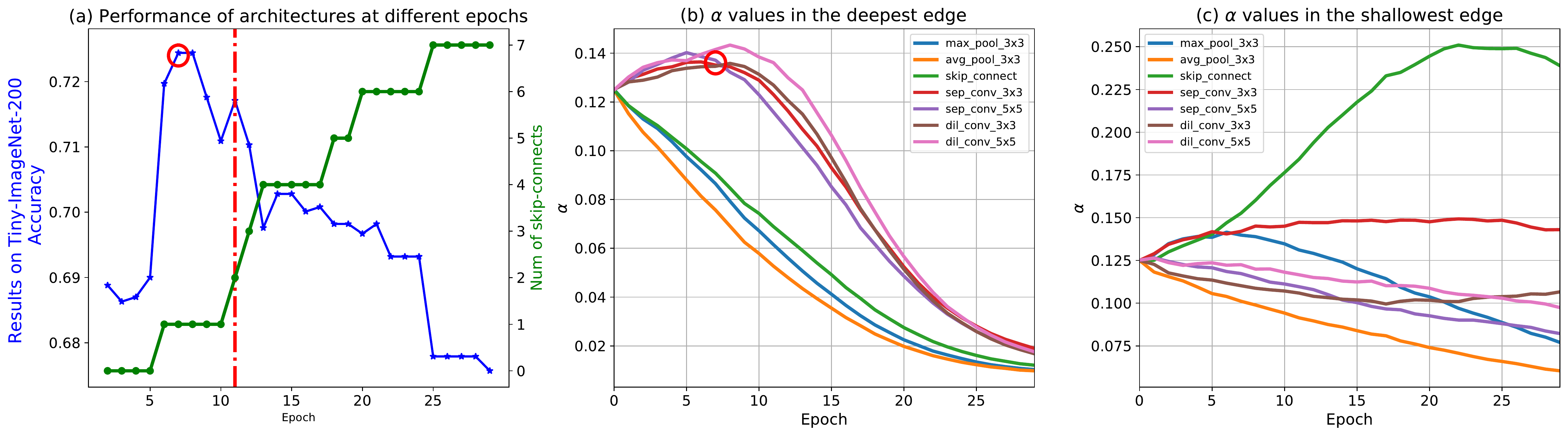}
        \caption{More results on Tiny-ImageNet-200 dataset. The meaning of each figure is the same as that in Fig. 2 in the main paper.}
        \label{all_in_three_200}
    \end{center}
\end{figure*}

\subsubsection{ResNet Search Space.}

Considering the success of ResNet~\cite{he2016deep} on various vision tasks, we use ResNet-50 as the backbone to develop the architecture search space. As the input size is $32 \times 32$, the architecture is modified such that the first layer is 3x3 convolution with stride 1, and the first max-pool layer is removed, so that the output feature map size before the global average pooling layer is $4 \times 4$. We construct the one-shot model by replacing {\em conv-3x3} in each residual block with a search cell. Similar to the inception module~\cite{szegedy2016rethinking}, a search cell is composed of 4 branches of candidate operations. The goal is to choose one operation per branch by applying the softmax operation within the branch. The outputs of each branch are concatenated together as the final output of the search cell. Note that every search branch involves 10 candidate operations including {\em zero, max-pool-3x3, avg-pool-3x3, sep-conv-3x3, sep-conv-5x5, sep-conv-7x7, dil-conv-3x3, conv-3x3, conv-5x5, conv-7x7}. Softmax is applied to architecture parameters to compute the weights, which are used to determine the selected architecture. We search the model on CIFAR100 for at most 80 epochs and use Criterion 2 for early stopping based on the ranking of the last layer's architecture parameters.

The selected architecture with early stopping Criterion 2 is shown in Fig. \ref{res_architect}. 

% which can be seen in Fig.~\ref{mvb2_final}. The search procedure is early stopped at epoch 47 (marked as the red circle in Fig. \ref{mvb2_final}, and we use the selected architecture at this epoch for evaluation. We also choose other architectures for fair comparison, including the architecture at epoch 30 and epoch 50. As shown in Table~\ref{tab:supp_result}, the architecture obtained from DARTS+ gets the best test accuracy of 81.23\%, conveying the effectiveness of the proposed early stopping scheme.

\begin{figure}[htbp]
    \centering
    \includegraphics[width=3in]{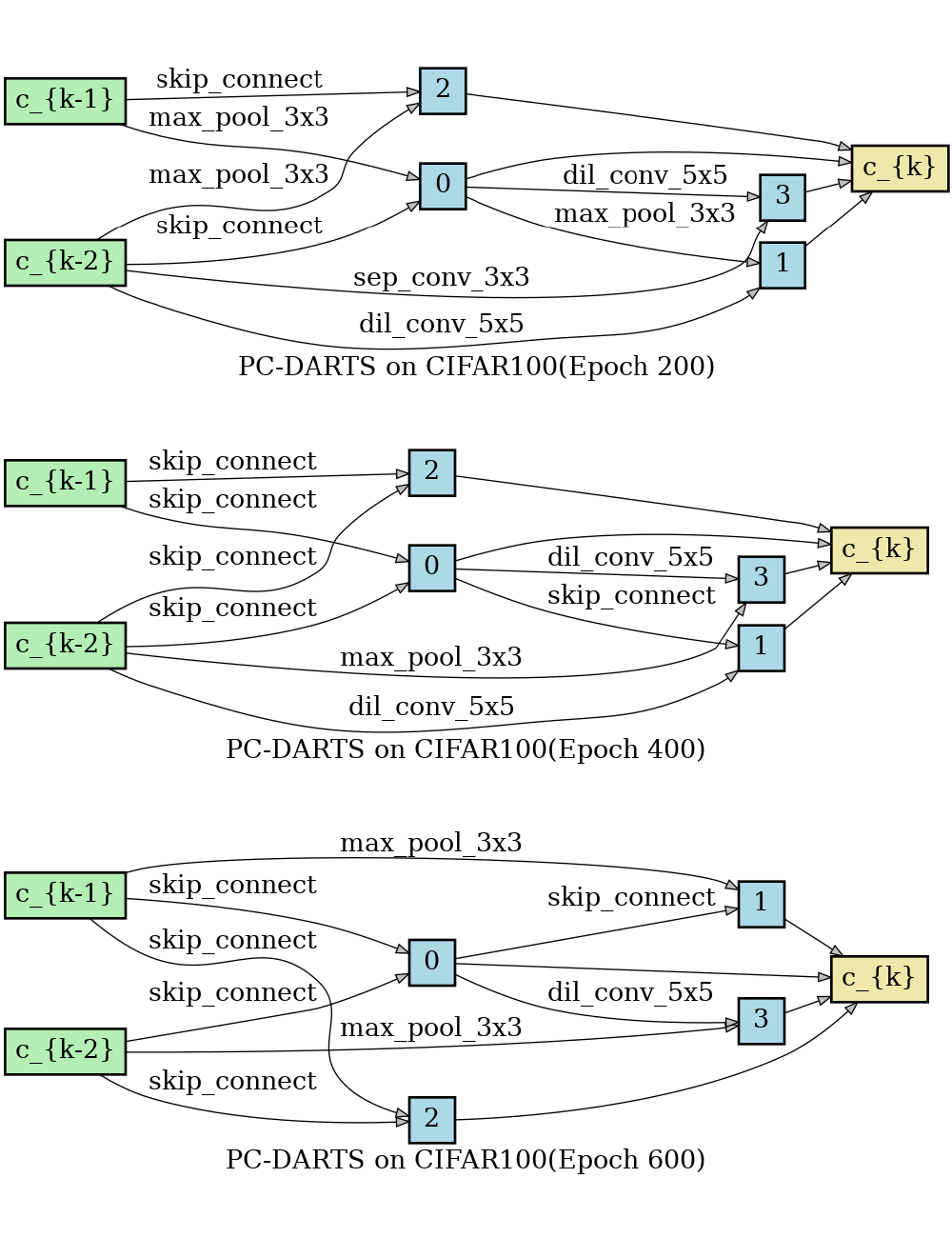}
    \caption{Selected normal cells at different epochs with PC-DARTS on CIFAR100.}
    \label{pc_darts}
\end{figure}

\subsection{Architecture Evaluation}

\subsubsection{DARTS Search Space.}
For each selected architecture, we follow the configurations and hyper-parameters of the previous works~\cite{liu2018darts,chen2019progressive} for evaluation on different datasets.

When evaluating CIFAR10 and CIFAR100, we use a network of 20 cells and 36 initial channels for evaluation to ensure a comparable model size as other baseline models. We use the whole training set to train the model for 2000 epochs with batch size 96 to ensure convergence. Other hyper-parameters are set the same as the ones in the search stage. Following existing works ~\cite{pham2018efficient,zoph2018learning,liu2018progressive,real2019regularized}, we also add some enhancements including cutout, path dropout with probability 0.2 and auxiliary towers with weight 0.4. We further increase the initial channel number from 36 to 50 and add more augmentation tricks including AutoAugment~\cite{cubuk2018autoaugment} and mixup~\cite{zhang2017mixup} to achieve better results. All the results are shown in Table 1 in the main paper.

For Tiny-ImageNet-200, the network is similar to CIFAR10/100 where 20 cells and 36 channels are involved, except that an additional $3 \times 3$ convolution layer with stride 2 is inserted in the first layer. We also transfer the architectures searched from other algorithms to Tiny-ImageNet-200 and evaluate the performance for a fair comparison. Other experimental settings are the same as CIFAR10/100. The results are shown in Table 2 in the main paper.

For ImageNet, We use the architecture searched directly from ImageNet for evaluation, and the architecture from CIFAR100 to test the transferability of the selected architecture. We follow DARTS such that the number of cells is 14, and the initial number of channels is 48. We train the model for 800 epochs with batch size 2048 on 8 Nvidia Tesla V100 GPUs as more epochs can achieve better convergence. The model is optimized with the SGD optimizer with an initial learning rate 0.8 (cosine decayed to 0), the momentum of 0.9, and weight decay $3\times10^{-5}$. We use learning rate warmup~\cite{goyal2017accurate} for the first 5 epochs and other training enhancements including label smoothing~\cite{szegedy2016rethinking} and auxiliary loss tower. We also adopt SE-module~\cite{hu2018squeeze} in the architecture transferred from CIFAR100, and introduce AutoAugment~\cite{cubuk2018autoaugment} and mixup~\cite{zhang2017mixup} for training to obtain a better model. The experimental results are shown in Table 3 in the main paper.

\subsubsection{MobileNetV2 and ResNet Search Space.}

Then, we follow almost the same configurations and hyper-parameters as used in DARTS~\cite{liu2018darts,chen2019progressive} to evaluate the selected architectures, except that the number of training epochs is 1,200 in MobileNetV2 and 600 in ResNet. To show the effectiveness of the selected architecture with early stopping, we train the architectures selected at other epochs for comparison. Fig. 4 in the main paper shows the effectiveness of the selected architecture with DARTS+, achieving 72.43\% in the MobileNetV2 search space and 81.23\% in the ResNet search space.

\section{Additional Experiments}

\subsection{More Illustrations on the Collapse of DARTS}\label{more-exps}

To further show the collapse problem in DARTS and validate the effectiveness of the proposed criteria, we perform more experiments on CIFAR10, CIFAR100, and Tiny-Imagenet-200 datasets. The experimental settings are the same as those in the main paper. For each selected cell at different epochs, we conduct extensive experiments for evaluation. The experimental results are shown in Fig.\ref{all_in_three_cifar10}, \ref{all_in_three_cifar100} and \ref{all_in_three_200}. We can arrive at similar conclusions shown in Sec. 2.2 in the main paper and both criteria are shown in Sec. 3 select similar architectures and the architectures achieve comparable performance.

\subsection{Implicit Early Stopping in PC-DARTS}\label{imp-es}

We carry out experiments to show that some recent progress of DARTS such as PC-DARTS~\cite{xu2019pc} adopts the early stopping idea implicitly. We use the original codes and settings of PC-DARTS~\cite{xu2019pc} and searching on CIFAR100 for 600 epochs. The selected normal cells at 200/400/600 epochs in the searching procedure are shown in Fig.\ref{pc_darts}. As the number of epoch increases, the number of skip-connects in the selected normal cells increases (2 skip-connects at 200 epochs and 5 at 600 epochs). It implies that PC-DARTS may suffer from collapse with more searching epochs, and training with just 50 epochs (used in DARTS and PC-DARTS) is an implicit early-stopping scheme to obtain better architecture.

\end{document}